\documentclass[a4paper,fleqn]{cas-sc}

\usepackage[authoryear,longnamesfirst]{natbib}

\usepackage{multirow, booktabs, tabularx, ragged2e}
\newcolumntype{P}[1]{>{\RaggedRight\hspace{0pt}}p{#1}}
\newcolumntype{L}[1]{>{\RaggedRight\hsize=#1\hsize\linewidth=\hsize}X}

\def\tsc#1{\csdef{#1}{\textsc{\lowercase{#1}}\xspace}}
\tsc{WGM}
\tsc{QE}
\tsc{EP}
\tsc{PMS}
\tsc{BEC}
\tsc{DE}
\begin{document}
\let\WriteBookmarks\relax
\def\floatpagepagefraction{1}
\def\textpagefraction{.001}

\shorttitle{Discovering robust biomarkers of psychiatric disorders from rs-fMRI via GNN: A systematic review}

\shortauthors{Chan et~al.}

\title [mode = title]{Discovering robust biomarkers of psychiatric disorders from resting-state functional MRI via graph neural networks: A systematic review}

\author[1]{Yi Hao Chan}[orcid=0000-0002-2393-1110]

\affiliation[1]{
    organization={College of Computing and Data Science, Nanyang Technological University}, 
    city={Singapore}, 
    postcode={639798}, 
    country={Singapore}
}

\author[1]{ Deepank Girish}

\author[2]{ Sukrit Gupta}
% \credit{Data curation, Writing - Original draft preparation}

\affiliation[2]{organization={Department of Computer Science and Engineering, Indian Institute of
Technology},
    city={Ropar},
    postcode={140001}, 
    state={Punjab},
    country={India}}

\author[1]{ Jing Xia}
\author[1]{ Chockalingam Kasi}
\author[1]{ Yinan He}
\author[1]{ Conghao Wang}

\author[1]{ Jagath C. Rajapakse}
\cormark[1]
\cortext[cor1]{Corresponding author}

\begin{abstract}
Graph neural networks (GNN) have emerged as a popular tool for modelling functional magnetic resonance imaging (fMRI) datasets. Many recent studies have reported significant improvements in disorder classification performance via more sophisticated GNN designs and highlighted salient features that could be potential biomarkers of the disorder. However, existing methods of evaluating their robustness are often limited to cross-referencing with existing literature, which is a subjective and inconsistent process.
In this review, we provide an overview of how GNN and model explainability techniques (specifically, feature attributors) have been applied to fMRI datasets for disorder prediction tasks, with an emphasis on evaluating the robustness of potential biomarkers produced for psychiatric disorders. Then, 65 studies using GNNs that reported potential fMRI biomarkers for psychiatric disorders (attention-deficit hyperactivity disorder, autism spectrum disorder, major depressive disorder, schizophrenia) published before 9 October 2024 were identified from 2 online databases (Scopus, PubMed).
We found that while most studies have performant models, salient features highlighted in these studies (as determined by feature attribution scores) vary greatly across studies on the same disorder. Reproducibility of biomarkers is only limited to a small subset at the level of regions and few transdiagnostic biomarkers were identified. To address these issues, we suggest establishing new standards that are based on objective evaluation metrics to determine the robustness of these potential biomarkers. We further highlight gaps in the existing literature and put together a prediction-attribution-evaluation framework that could set the foundations for future research on discovering robust biomarkers of psychiatric disorders via GNNs.
\end{abstract}

\begin{graphicalabstract}
\includegraphics[width=\textwidth]{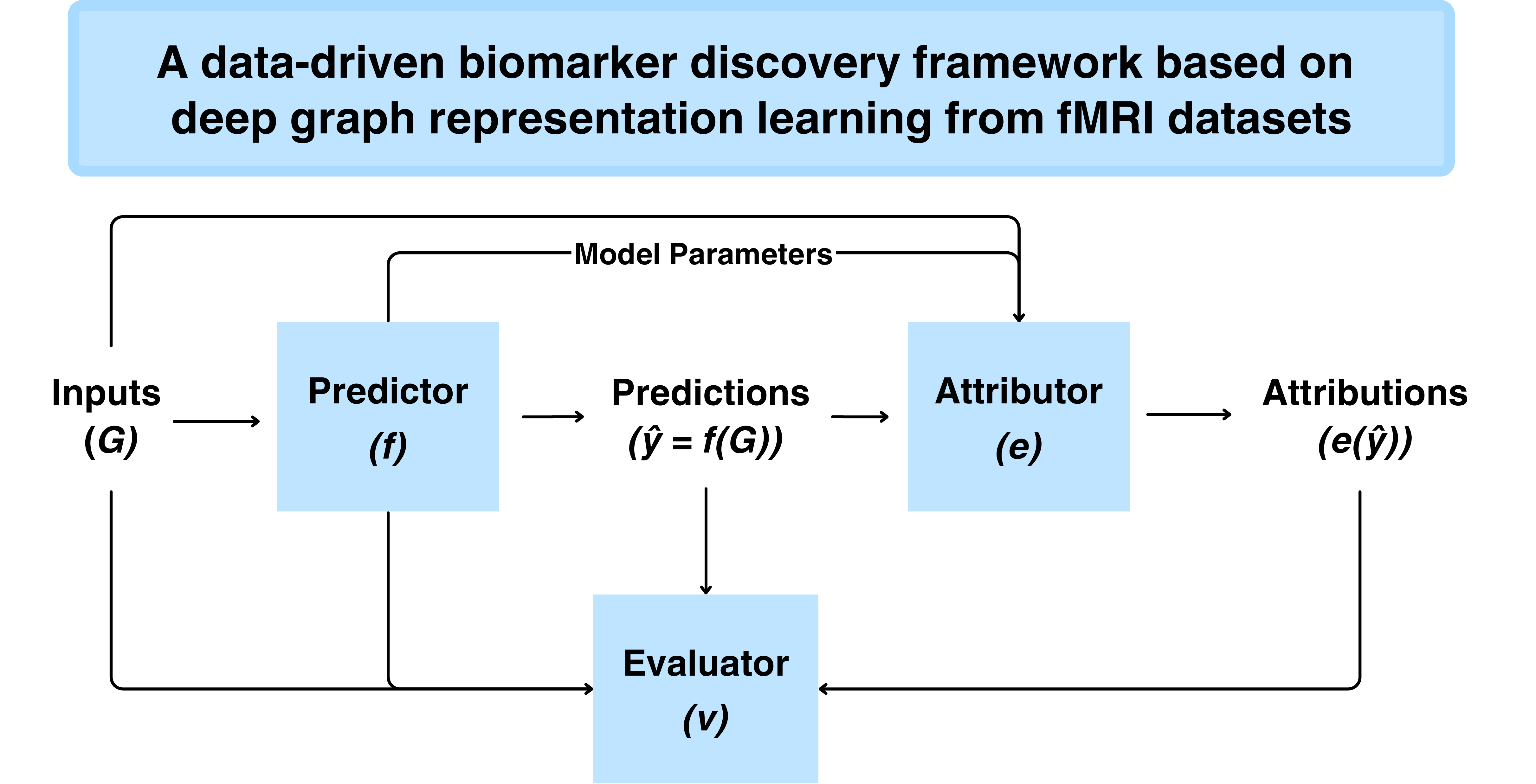}
\end{graphicalabstract}

\begin{highlights}
\item A wide range of predictors and attributors have been used in existing fMRI studies.
\item Robustness of salient features highlighted by attributors remains underexplored.
\item We propose a taxonomy to classify state-of-the-art GNNs customised to fMRI datasets. 
\item Some key features at the level of regions of interest were consistent across studies.
\item More evaluation metrics are needed to demonstrate robustness of potential biomarkers.
\end{highlights}

% Keywords

\maketitle

\section{Introduction}\label{sec1}

Psychiatric disorders often manifest as changes in the functional characteristics of the brain. Functional magnetic resonance imaging (fMRI) has been widely used to objectively quantify these functional aberrations and identify the underlying neural substrates in the human brain. 
This has led to decades of research documenting the associations between psychiatric disorders and disruptions in whole-brain functional connectivity (FC) \citep{filippi2023human}. However, these characterisations have yet to reveal strong and reproducible biomarkers for most disorders \citep{abi2023candidate,chollet2022functional}. 
The lack of success is commonly attributed to limitations such as 
disease heterogeneity \citep{verdi2021beyond},
inter-individual variability \citep{canario2021review},
small effect size \citep{poldrack2017scanning,jia2018small}, 
noise \citep{liu2016noise},
limited dataset sizes \citep{marek2022reproducible},
site effects (also known as batch effects) when combining data from multiple sites \citep{bayer2022site} and
variability introduced by the choice of pre-processing pipeline \citep{dadi2019benchmarking,botvinik2020variability}. 

Over the past years, larger datasets have emerged through inter-institution collaborations \citep{laird2021large} and standardised formats for organising neuroimaging datasets \citep{gorgolewski2016brain}, more mature pre-processing pipelines are available (for example, fMRIPrep \citep{esteban2019fmriprep}), and better harmonisation tools have been developed to reduce batch effects \citep{hu2023image}, alleviating some of the above-mentioned issues in fMRI studies. Coupled with the development of more sophisticated modelling tools, model performances have improved and more potential biomarkers have been discovered in recent years, warranting the need for another review to synthesise their findings.

Many modern modelling tools involve machine learning (ML) algorithms, partly due to the high dimensionality of fMRI datasets. 
Mass univariate approaches (where each voxel is modelled independently) have traditionally been used, but they do not capture inter-region functional relationships \citep{habeck2010multivariate}. 
To address this limitation, multivariate techniques (e.g. multi-voxel pattern analysis \citep{weaverdyck2020tools}) have been proposed, involving ML algorithms such as support vector machines (SVM). 
More recently, deep learning models have been shown to outperform SVM in disease classification tasks \citep{zhang2020survey}. While convolutional neural networks (CNN) customised for connectome datasets \citep{kawahara2017brainnetcnn,meszlenyi2017resting} were proposed as an improvement over vanilla deep neural networks (DNN) \citep{gupta2021obtaining}, graph neural networks (GNN) have since emerged as the state-of-the-art deep learning model used in network neuroscience studies \citep{bessadok2022graph}. Besides being an intuitive fit to FC matrices (which are best represented as graphs), the flexibility afforded by GNNs makes it possible to design techniques that capture intra-modular relationships \citep{wang2024leveraging} or even inter-patient relationships \citep{parisot2018disease}. While vanilla GNNs do not seem to do better than DNNs and CNNs \citep{elgazzar2022benchmarking}, more carefully designed GNN architectures have demonstrated significant improvements in disease classification performance \citep{song2021graph,xiao2022dual}. 

However, disorder prediction is rarely the end goal as clinical adoption of such models is rare \citep{duda2023reliability}. Instead, these models can be used to provide neurological insights (e.g. nosology, subtyping, etc.). This is most commonly demonstrated via model explainability techniques (henceforth termed as `attributor', as the majority of them assign an importance score to each feature as a limited form of explanation), ranging from gradient-based methods such as Integrated Gradients (IG) \citep{chan2022semi} or perturbation-based approaches such as GNNExplainer \citep{gallo2023functional}. 
Alternatively, GNN-specific mechanisms such as graph pooling \citep{li2021braingnn} can simultaneously train the model and produce feature attribution scores.
These attributors typically assign attribution scores to each feature or identify important subgraphs that contribute most to the model's predictions.  

While many attributors have been used in existing studies, several GNN-specific attributors remain unexplored. Furthermore, little has been studied about the robustness of these attributions. 
Existing studies often provide a very limited evaluation of their biomarkers, only reporting the top few features and cross-referencing other studies that had similar findings. Most recently, several studies have started comparing the attributions generated by different feature attributors and found that they could vary across datasets \citep{gallo2023functional}, predictors \citep{zhang2022identifying} and even attributors even when the same predictor was used \citep{hu2021gat,li2023classification}.
To ensure that salient features are truly representative of disorder traits and not mere artefacts, it would be prudent to take a pause and survey the existing literature to identify any convergence in the potential biomarkers that they reported.

In this review, we summarize recent progress on psychiatric disorder prediction via GNNs, with a focus on reviewing attributors used and potential biomarkers discovered by these fMRI studies. Psychiatric disorders included in this review include 
attention-deficit hyperactivity disorder (ADHD), 
autism spectrum disorder (ASD), 
major depressive disorder (MDD) and schizophrenia (SZ).
We explore the following questions, highlight remaining research gaps in these areas, and suggest pathways for future research on biomarker discovery via GNNs:
\begin{enumerate}
    \item Are there any graph construction approach and GNN-specific architecture designs that have consistently demonstrated superiority over others (e.g. their proposed GNN-based model works well across multiple datasets and studies)?
    \item How do existing studies evaluate the robustness of the potential biomarkers identified by their proposed approach (i.e. combination of predictor and attributor)?
    \item For each disorder, is there any convergence of discovered potential biomarkers across multiple studies (that are based on machine learning techniques)? If there are any, are they present in other disorders as well (i.e. potential transdiagnostic biomarkers)?
\end{enumerate}

\subsection{Related studies}
\label{sec:related}

Several review papers have been written on topics such as the use of GNNs for fMRI data analysis, model explainability in GNNs, and biomarker discovery for various neurodegenerative diseases and neuropsychiatric disorders. However, none has attempted to study these topics in a cohesive manner.
Recent reviews on GNN applications in network neuroscience \citep{bessadok2022graph,zhang2023combination} summarised various graph-based and population-based models that have been proposed.
Several benchmarking studies \citep{elgazzar2022benchmarking,cui2022braingb,said2023neurograph} have also been conducted to analyse the efficacy of various GNN designs on fMRI data in a fair and controlled manner that minimises the effect of covariates. 
Our study builds on top of their findings by synthesizing their key takeaways and goes beyond the brain graph - population graph dichotomy to provide a taxonomy of state-of-the-art GNN architectures that are customised for fMRI datasets. 

Model explainability methods have been thoroughly reviewed in previous works \citep{linardatos2020explainable,tjoa2020survey,marcinkevivcs2023interpretable}, including methods that are specialised for GNNs \citep{yuan2022explainability,kakkad2023survey}. Several recent reviews have examined the use of these methods in the medical domain \citep{munroe2024applications,rahman2023looking}.
However, to the best of our knowledge, no review has been done to assess the relevance of these methods in the context of biomarker discovery from fMRI datasets. 
Existing reviews on biomarkers \citep{abi2023candidate,filippi2023human} tend to summarise various types of biomarkers (often going beyond fMRI) and do not focus on examining and evaluating the reliability of the computational techniques \citep{nauta2023anecdotal,agarwal2023evaluating} used to derive the biomarkers. 
In our review paper, we aim to address this gap.

\subsection{Review methodology}
\label{sec:method}

\begin{figure*}
	\centering
		\includegraphics[width=\textwidth]{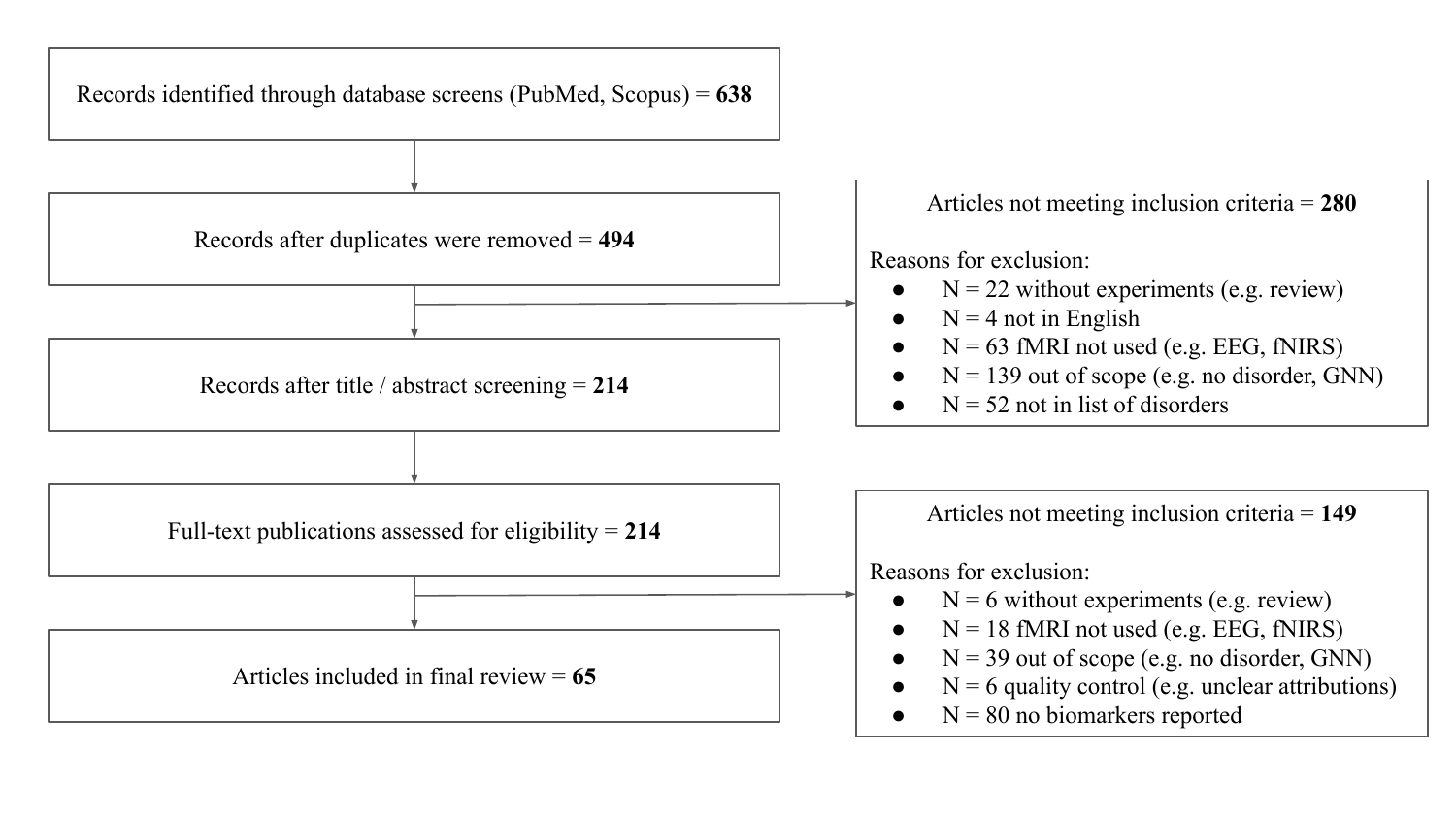}
	\caption{Flowchart detailing the selection process of this review.}
	\label{fig:prisma}
\end{figure*}

To identify papers that applied GNNs on fMRI datasets, we performed a search on 9 October 2024 via PubMed and Scopus with the following search query: (``graph neural networks" OR ``graph convolutional networks" OR `` GNN " OR `` GCN ") AND (``fmri" OR ``functional MRI" OR ``functional connectivity"). 
158 papers from PubMed and 480 papers from Scopus matched these search terms.
Manual filtering was conducted to remove duplicates and remove irrelevant results (e.g. no disorder prediction, resting-state fMRI not used in experiments, no experiments conducted). 
For papers on disorder prediction, only papers within our defined scope (ADHD, ASD, MDD, SZ) were included and they should report biomarkers clearly (e.g. not just reporting a brain map without any labels) as one major goal of our paper is to consolidate and identify potential biomarkers reported across multiple studies. This was checked manually and independently by 3 reviewers who scanned through the full text.
Figure \ref{fig:prisma} shows a detailed breakdown of the filtering process. 
At the end of this filtering process, a total of 65 papers were reviewed. 
The review and its protocol were not registered as retrospective registration is not feasible. Nevertheless, the PRISMA checklist \citep{page2021prisma} and review protocol can be found in the Supplementary materials.
Data such as the number of subjects, class splits (disorder, controls), type of GNN, model performance (accuracy), biomarkers reported, etc. were manually extracted from these papers. 
Variations in labelling techniques (i.e. criteria for diagnosis) across studies were present but minimal as a majority of the studies, for each disorder, relied on the same dataset source. Notably, such variations are more pronounced within each study as datasets are often aggregated from multiple imaging sites which often do not have exactly the same inclusion/exclusion criteria. As an exploratory review, studies were not excluded on the basis of having different labelling techniques. 
In occasional cases of missing information and there was no feasible way to retrieve them (e.g. model accuracy presented in charts but numbers not reported), these fields were left blank in the tables reported below and not considered in the computation of mean statistics (e.g. mean accuracy). However, studies without any biomarkers clearly reported were excluded.
Several studies tested their models across multiple diseases. In such cases, data specific to the disorder (e.g. dataset, class splits) are extracted separated for each study. 
Potential biomarkers reported were manually collated and matched across studies as there is no standardised reporting format, nor any existing tools that could automate the process.

\section{Modelling functional MRI datasets for disorder prediction}
\label{sec:modelling}

Blood-oxygen-level-dependent signals captured in fMRI scans are fundamentally represented as time series data from individual voxels. Even at relatively low resolutions (e.g. 5mm), the number of voxels ($>10,000$) far outnumbers typical dataset sizes. Coupled with the issue of low signal-to-noise ratio (SNR) in fMRI data \citep{vizioli2021lowering}, these problems have motivated researchers to group clusters of related voxels together to improve SNR. Examples of such techniques are atlas-based approaches, independent component analysis (ICA) \citep{calhoun2009review}, and functional gradients/manifolds \citep{hong2020toward}. The former applies atlases developed by delineating boundaries following anatomical landmarks (such as sulci \citep{rolls2020automated} and gyri \citep{desikan2006automated}) or task-fMRI experiments that identify regions of interest (ROI) that are activated when performing various tasks \citep{seitzman2020set}.
This provides an informed way of feature selection to reduce data dimensionality, with the disadvantage of neglecting the voxels that are not part of the atlas' ROIs if a sphere-based approach (i.e. demarcate voxels that are within a certain radius from the ROI's coordinate) is used instead of a parcellation-based approach. 
On the other hand, ICA and functional gradients take a different approach by learning lower-dimensional representations of the original data. 
Out of these approaches, atlas-driven dimensionality reduction techniques are most widely used in the studies considered in this review.

\subsection{Functional connectivity}

Besides modelling the mean time series of each ROI or component directly, another common way to analyse fMRI data is to study the relationship between pairs of time series data. 
Pearson correlation has been the most common approach to compute such FC matrices. However, it is limited to capturing linear correlations and it could have weaker connections suppressed by noise or imaging artifacts. Alternative metrics introduced to address these limitations include various forms of partial correlation and sparse representation \citep{yang2021autism}. 

While a majority of existing studies are limited to such pairwise analysis at the level of regions/nodes, several recent studies have experimented with new graph construction techniques to address existing limitations.
Going beyond inter-nodal relationships, edge FC was proposed to consider relationships between edges \citep{faskowitz2020edge} (via much larger FC matrices, as all pairwise combinations of nodes are considered). Going beyond pairwise relationships, hypergraphs are higher-order graphs where each hyperedge represents the relationship between two or more nodes. In recent studies, hypergraphs based on connectomes are typically generated dynamically \citep{wang2023dynamic,ji2022fc}.

Regardless of the choice of analysis type (i.e. region-level or edge-level pairwise connectivity, or higher-order relationships), there are two key paradigms of modelling FC: static FC (sFC) and dynamic FC (dFC). sFC assumes that FC is stable throughout the scan (i.e. Pearson's correlation is computed from the entire time series, without splitting it into parts). On the other hand, dFC does not make such an assumption. The most common approach involves using a sliding window across the time series, generating multiple FC matrices in the process. More sophisticated approaches perform clustering and decomposition on these matrices to produce a single dFC matrix that better captures FC dynamics than the sFC matrix.
Du \emph{et al.} \citep{du2018classification} provides a detailed review of dFC (as well as ICA-based methods).
Overall, using sFC at the level of pairwise ROIs remains the most common approach despite the above-mentioned advancements.

\subsection{Graph representations of fMRI datasets}

\begin{figure*}
	\centering
		\includegraphics[width=\textwidth]{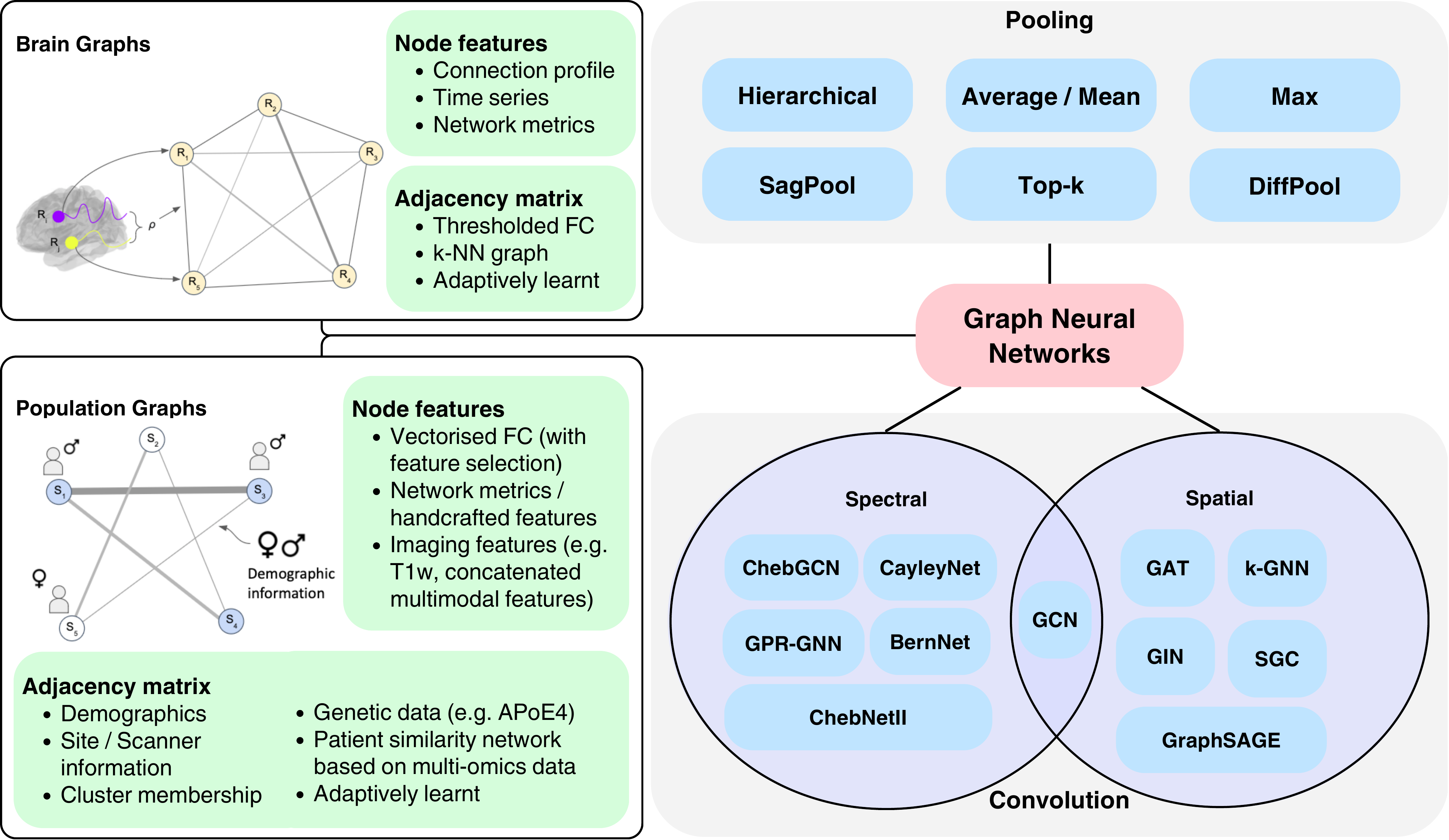}
	\caption{Summary of the key components of typical GNNs used on fMRI datasets.}
	\label{fig:predictors}
\end{figure*}

fMRI datasets, especially FC matrices, are most naturally represented as graphs. A graph data structure consists of nodes that are linked by edges, which represent the relationship between connected nodes. There are two major paradigms of modelling fMRI datasets: brain graphs (BG) and population graphs (PG). Figure \ref{fig:predictors} illustrates the differences between them and lists down several possible options when choosing the node features and adjacency matrix of the BG and PG.

In a BG, each node in the graph represents an ROI and the choice of node features varies across studies - most commonly, the connection profile for that ROI (i.e. the row in the FC matrix that corresponds to that ROI) is used. 
Other possible node features include regional features that compute various statistics of the voxels within an ROI, such as regional homogeneity (ReHo) and Amplitude of Low Frequency Fluctuations (ALFF) \citep{arya2020fusing}.
Edges store quantitative measures of the relationship between ROI pairs (e.g. Pearson's correlation of the mean time series). 
Thus, each subject (or scan) is represented as a graph, and graph classification is typically performed.
On the other hand, in a PG, each node represents a subject (or scan) and the graph represents the population of interest. Node classification is typically performed.
Node features typically store some representation of the imaging data (often after dimensionality reduction via recursive feature elimination (RFE) or principal component analysis (PCA)).
Edges store measures of similarity between scans (usually demographic information such as age and gender, or any vector from which distance can be computed). Another alternative is to construct a k-nearest neighbours (k-NN) graph \citep{wang2021graph,zhang2021graph}. 

PGs allow a much wider variety of data (including metadata) to be incorporated into the analysis. However, such graphs are typically pre-defined rather arbitrarily and tend to be static \citep{bintsi2023comparative}. Thus, learnable \citep{cosmo2020latent} and adaptive \citep{song2021graph,park2023graph} methods of PG construction have been proposed.  
Recent studies \citep{jiang2020hi,xiao2022dual,zhang2022classification,he2023predicting} have also explored ways to use BG and PG simultaneously. 
Overall, BG is the dominant approach of input graph construction (used in 82.1\% of studies, as compared to 7.7\% for PG). Connection profile is by far the most popular node feature (58.9\%) and the adjacency matrix is often the thresholded FC matrix (47.4\%).

\subsection{Encoding fMRI data with graph neural networks}

Let $G = (V, A, X)$ represent a graph used by the GNN, where $V$ represents the set of nodes in the graph, $A \in \mathbb{R}^{|V| \times |V|}$ represents the adjacency matrix used and $X \in \mathbb{R}^{|V| \times K}$ represents the node features, each of length $K$.

Traditionally, network-based analyses have been performed on fMRI datasets for disease studies, revealing insights such as lower clustering coefficient, global efficiency, and node degree for patients with mild cognitive impairment \citep{filippi2023human}.
With the advent of ML, researchers started training models that distinguish between healthy subjects and patients. Since graphs cannot be used as input to many of these models, features were handcrafted (e.g. using network metrics \citep{yin2021graph}) or in the case of FC matrices, vectorised by flattening the lower triangular \citep{gupta2021obtaining}. Doing so loses the graph structure and leads to a high-dimensional input. Thus, models tend to overfit and feature selection methods (such as two-sample t-test and recursive feature elimination) are often used to address these issues \citep{teng2023brain}.

Recently, GNNs - neural networks that are designed to be applied directly to graphs - have been used for encoding fMRI datasets. GNNs provide a parameter efficient means of modelling FC matrices \citep{li2021braingnn}, reducing the problem of overfitting and allowing for more sophisticated analyses involving modular brain networks \citep{mei2022modular} and multimodal data \citep{he2023predicting}. GNNs can be broadly categorized into spectral GNNs and spatial GNNs. 

Spectral GNNs perform convolution by transforming the graph signal and filtering to the spectral domain before convolving. The convolution operation is defined as:

\begin{equation}
    g \star x = U(U^T g \circledast U^T x),
\end{equation}

where $U^T x$ converts a signal $x$ to the spectral domain using graph Fourier transform and $U x$ transforms the signal $x$ back to the spatial domain using inverse graph Fourier transform. The operation can be simplified as:

\begin{equation}
    g_{\theta} \star x = U g_{\theta} U^T x,
\end{equation}

where $g_{\theta}$ denotes a learnable diagonal matrix.
Examples of spectral GNN include ChebNet \citep{defferrard2016convolutional} and a recent improvement of it called ChebNetII \citep{he2022convolutional} which noted that ChebNet can learn inappropriate Chebyshev coefficient which results in overfitting and sub-optimal performance. ChebNetII uses Chebyshev interpolation to overcome this issue, demonstrating better performance. 

Spatial GNNs, on the other hand, apply convolutions to the graph based on its topology. Adopting a message-passing paradigm, node features in node $i$ are iteratively updated by aggregating node features from its neighbours $\mathcal{N}(i)$.

\begin{equation} 
    X'_i = \alpha_{i,i} X_i U + \sum_{j \in \mathcal{N}(i)} \alpha_{i,j} X_j W,
\label{eq:gnn}
\end{equation}

where $U, W \in \mathbb{R}^{K \times K'}$ represent learnable weights and $\alpha$ represents coefficients learnt to balance between retaining information from the original node vector and using information gathered from its neighbours.
A point to note is that the above formulations present the core convolution operations (without the non-linearity shown) but in actual implementations, normalisation of the adjacency matrix is often performed to prevent the problem of vanishing or exploding gradients when multiple graph convolution layers are used together. 

In general, variants of spatial GNN differ in how they weigh and aggregate information.
For example, graph attention network (GAT) \citep{velivckovic2017graph} learns attention scores to choose which neighbours to focus on, while graph isomorphism network (GIN) \citep{xu2018powerful} learns to balance between information from neighbours and the node's own node features.
Note that graph convolutional network (GCN) \citep{kipf2016semi} can be seen as both spatial GNN and spectral GNN, as shown in Figure \ref{fig:predictors}. 

Beyond baseline GNNs, several graph convolution layers and pooling layers have been proposed for use in connectome datasets. 
Many baseline GNNs focus on node-level aggregation, but FC is often used as the graph of BGs. This motivates the use of edge features as well as edge-based convolutions such as EdgeConv \citep{wang2019dynamic} or GraphConv \citep{morris2019weisfeiler}. 
Figure \ref{fig:predictors} also provides a summary of common baseline pooling layers. More details about pooling layers will be discussed in Section \ref{sec-pooling} as they are often used as a means to improve model explainability.

Additionally, Figure S\ref{fig:pie_pred} summarises the types of GCNs used in these studies. 
It is evident that GCNs are widely used (present in almost half of the studies) and spatial GCNs (e.g. GIN, GAT, GraphSAGE) are more widely used than spectral GCNs (e.g. ChebNet)

Details about the above GNN architectures have been reviewed by \citep{bessadok2022graph} and several benchmarking studies have evaluated the efficacy of GNNs on fMRI datasets. Thus, we will summarise key points from these papers and focus on aspects pertaining to model explainability.
BrainGB \citep{cui2022braingb} splits the GNN methodology into 4 parts: node feature construction, messaging passing mechanism, attention-enhanced message passing, and pooling strategies. 
On multiple datasets (both healthy subjects and patients with disorders), they showed that connection profile (i.e. use the corresponding row of the FC matrix as the node features for an ROI), node concat message passing with attention (i.e. multiply learnt attention weights to the neighbour's node feature before concatenating it with the node's feature vector, followed by a multi-layer perceptron), and concat pooling (i.e. concatenate node features from multiple ROIs when performing graph pooling) works best.

Neurograph \citep{said2023neurograph} applied various spectral and spatial GNNs (GCN, GAT, GIN, etc.) on healthy subjects from the Human Connectome Project dataset for various baseline problems such as task classification, gender classification and age prediction.
They showed that their proposed GNN$\star$ architecture (which has 3 graph convolution layers with skip connections) works best. 
They also experimented with various settings such as the number of ROIs used, the sparsity of graphs, and how the node features were created. Model performance was shown to improve when including more ROIs (400 and 1000), using sparser graphs (5\%), and using Pearson correlation for node features.
It is notable that another benchmarking study \citep{elgazzar2022benchmarking} suggested that GNNs do not even outperform 1D CNN for MDD and ASD classification. However, they opted for a relatively dense FC matrix (50-90\%) and binarised the graph.

Overall, all three benchmarking papers noted that the sparsity of the graph used by the GNN impacts model performance and lower sparsities (below 50\%) were shown to be beneficial. Hyperparameter and GNN construction choices that lead to more model parameters (e.g. more ROIs, concatenation of node features) seem to enhance performance, but this should be done with care even though GNNs are parameter efficient (relative to other DNNs). Finally, vanilla GNNs do not seem to clearly outperform non-graph baselines such as SVM and CNN, but more carefully designed GNNs do improve model performance.

\subsection{State-of-the-art GNN architectures customised for fMRI datasets}

\begin{figure*}
	\centering
		\includegraphics[width=\textwidth]{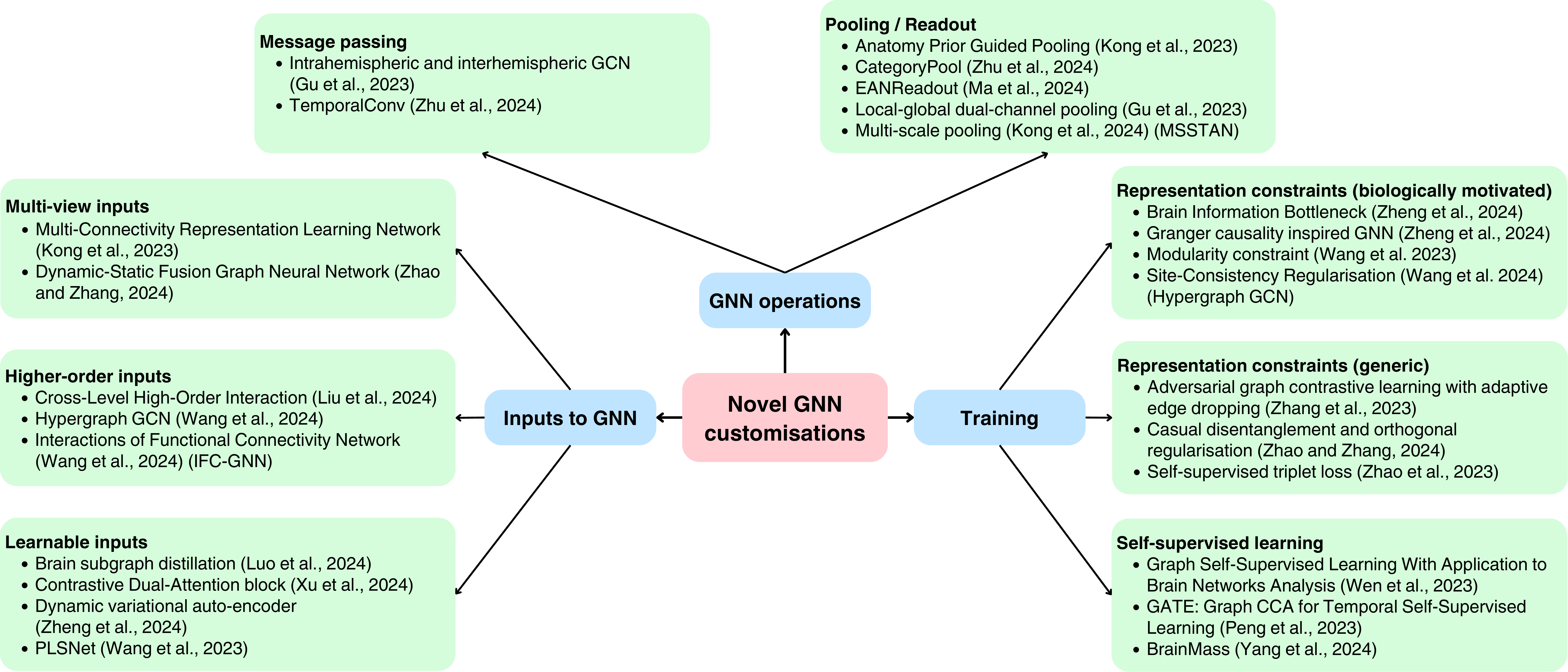}
	\caption{Our proposed taxonomy of state-of-the-art GNN models customised for fMRI datasets.}
	\label{fig:sota_pred}
\end{figure*}

The benchmarking studies discussed above largely involve baseline GNNs and they have not considered state-of-the-art GNN architectures that are customised for fMRI datasets. Such improved architectures have been shown to outperform baselines and generalise better than them. Thus, it would be of interest to use them (instead of baseline models) for biomarker discovery.

These models are often built on top of baseline GNN models, addressing certain limitations by making changes in the (i) input to the GNN, i.e. graph construction process, (ii) formulation of the GNN's message passing and pooling mechanism, (iii) techniques used to train these GNNs. Figure \ref{fig:sota_pred} shows our proposed taxonomy to categorise these customisations of the original baseline GNN. 

Most routine applications of GNN on fMRI datasets involve the use of vanilla GNNs (e.g. GCN, ChebNet) constructed using a binarised / thresholded FC matrix as the adjacency matrix and the connection profile as the node features. Numerous state-of-the-art GNN models go beyond this by introducing (i) techniques to learn representations of the original data and use them for graph construction, (ii) higher-order information (such as hypergraphs) to go beyond what a typical FC matrix can represent, and (iii) ways to incorporate multiple views (e.g. sFC and dFC, multiple atlases, etc.). Notably, in the case of hypergraphs, its higher-order relationships require a slightly different GNN formulation (which is actually a generalisation of GNNs \citep{bai2021hypergraph}). Instead of an adjacency matrix, an incidence matrix $B \in \mathbf{R}^{|V| \times |H|}$ is used, where $H$ represents the set of hyperedges. Each hyperedge $h \in H$ comes along with a diagonal weight matrix $W' \in \mathbf{R}^{|H| \times |H|}$. Another two diagonal matrices are also required: $D_{ii} \in \mathbf{R}^{|V| \times |V|}$ representing vertex degree and $D_{hh} \in \mathbf{R}^{|H| \times |H|}$ representing hyperedge degree. The message passing mechanism in hypergraph GCNs (HGCN) can then be expressed as: 

\begin{equation} 
    X'_i = \alpha_{iih} X_i H_{ih} H_{ih} W'_{hh} X_i U + \sum_{j \in {V\setminus\{i\}}} \sum_{h=1}^{H} \alpha_{ijh} H_{ih} H_{jh} W'_{hh} X_j W.     
\label{eq:hgnn}
\end{equation}

Another paradigm of state-of-the-art GNNs modifies the standard message passing and pooling operations so that the model conforms to pre-existing biological knowledge. For instance, instead of aggregating information from all neighbours or relying on attention to learn which neighbours to focus on, one could limit the scope to intra-hemispheric neighbours before inter-hemispheric aggregation \citep{gu2025fc}. Pooling is another active area with numerous innovations that make use of our current understanding of the brain to consolidate information across nodes in a biologically meaningful manner (rather than a coarse, one-step pooling operation) \citep{zhu2024temporal}.

A different class of modifications involves the introduction of constraints in the loss function used to train the GNN. 
Vanilla GNNs are typically only optimised with respect to the task (e.g. minimise cross-entropy). This gives the neural network too much freedom and it could end up learning representations that are not biologically sound. 
Adding constraints to the loss function could prevent this. 
For instance, a modularity constraint \citep{wang2023modularity} was introduced to ensure that embeddings learnt by nodes (ROIs) that originated from the same module were similar, in line with empirical observations of the modular architecture of the functional brain network. 
Several other studies focus on more generic constraints that aim to improve the robustness of the GNN to minor alterations of the FC matrix. This could have downstream implications on the robustness of biomarkers and would be an interesting area for further research.
Finally, self-supervised learning techniques have also been proposed to alleviate the problem of small fMRI datasets. Such techniques collate large datasets for pre-training GNNs, often relying on various novel data augmentation strategies \citep{peng2022gate,yang2024brainmass}. Such advancements enable the construction of foundation models, which have been shown to improve generalisation capabilities.

\subsection{Evaluation of strengths and weaknesses}

Overall, recent research on using GNNs on fMRI datasets for disorder prediction has experimented with a large variety of modelling techniques and proposed numerous customisations on top of baseline models. 
Thus, it would be valuable to evaluate them and suggest future research directions to move the field closer to the goal of producing more robust functional neuromarkers of psychiatric disorders. 
To keep the evaluation concise, the focus is kept on the best performing model configurations. These are identified by analysing studies with over 100 data samples and their disorder classification performance. Despite the limitations of classification accuracy as a metric (e.g. class imbalance), it was used as that was the most widely available metric and we found that the problem of class imbalance was not very prevalent in the selected studies. Overall, the mean accuracy reported was 75.7\%.

Answering the first question mentioned in the introduction (whether any particular graph construction approach and GNN architecture designs are better), the best-performing models typically used thresholded FC matrices as the adjacency matrix (i.e. retaining the values above the threshold) and connection profile as node features, concurring with the findings in \citep{cui2022braingb}. However, numerous studies with lower performance also used connection profiles but tend to involve binarised adjacency matrices. 
The simplicity of this predominant approach, coupled with its high performance, could explain why more advanced methods (e.g. sparse representation, learnable inputs, etc.) have not caught on. 
Furthermore, in biomarker discovery applications, the current goal is typically to cast a wide net across the whole brain and compute attributions relative to a complete set of features. Thus, sparse representation and learnable inputs might be less suitable in such contexts. Nevertheless, we note that sparsity (as shown in the benchmarking studies) and learning-based approaches are key features in the construction of the adjacency matrix, while preserving the original state of the node features for computing attributions.

Studies that used a mix of both BG and PG tend to perform better than solely using either graphs. Such an approach delivers the best of both worlds (BG allows efficient encoding of input data while incorporating information from the adjacency matrix, while PG allows meta-data such as demographic information to be incorporated easily). However, such an approach inevitably introduces more trainable parameters to the model and without proper benchmarking done, it is uncertain whether the improvements reported are simply a result of larger model capacity (i.e. if the BG-only or PG-only model were to be scaled up to the same number of parameters, they might close up the performance gap).

One major issue with the predominant approach is the choice of brain atlases: in particular, the AAL atlas is a popular choice, being used in 58\% of the studies.
While the consistent use of the same atlas across studies has the advantage of reducing one source of variability for future meta-studies and reviews, the suitability of the choice of atlas for the target population is often a point of contention. AAL, being derived from a single `healthy' young adult, might not be the best choice for studies on adolescents or children (e.g. for ASD, ADHD). Additionally, the versions most frequently used (AAL90 and AAL116) have fewer parcels than other atlases, resulting in large ROIs that provide less value as they represent sizable areas of the brain that would be too heterogeneous for any meaningful insight to be drawn. 
Some studies have proposed multi-atlas strategies, but it is still unclear how best to reconcile disagreements in attributions across atlases (e.g. if there are overlaps in ROIs/voxels across atlases, but they receive very different attributions). Thus, it would be necessary for such issues to be addressed, or alternative dimensionality reduction techniques (e.g. ICA, functional gradients) could be more thoroughly analysed in future studies.

In terms of the GNN construction, neither spatial nor spectral GCN has a clear lead over the other in terms of model performance. However, out of the classes of GNN architectures, the pooling-based approach featured more frequently among the best-performing models.
Most of these advancements provide a clear value over baseline GNNs in terms of their utility for biomarker discovery as the unconstrained nature of baseline GNNs opens it up to the problem of learning spurious relationships.
However, many studies did not control nor report the model size (i.e. number of parameters) and do not always compare with the latest models (in part, due to the different focus each model typically has). Thus, it would be prudent to rely on new benchmarking studies, with state-of-the-art GNNs considered, that control for the model size before coming to any conclusions. 
However, even if significant improvement in model performance is introduced by these novel techniques, another crucial consideration is the impact on the model attributions.
For instance, the impact of using synthetic data for graph augmentation in self-supervised pre-training has not been thoroughly evaluated for biomarker discovery purposes (e.g. Does the introduction of synthetic data lead to spurious biomarkers? Do the additional datasets used for augmentation drown out actual signals from the original dataset of interest?).

\section{Computing feature attributions via model explainability techniques}

Although machine learning models trained on fMRI data can perform disorder classification well, they are rarely adopted clinically \citep{duda2023reliability}. Thus, studies involving such models often provide additional insights via various model explainability approaches. One type of explanation involves deriving feature attributions from a model trained on disorder classification tasks. These attributions represent the importance of each feature. Features with high attribution scores could be potential biomarkers of the disorder, i.e. identifying characteristics of the disorders (e.g. particularly low connectivity between 2 ROIs, across the population with the disorder and not present in the healthy population).

Early studies analyzed neuroimaging data to study disorders using univariate \citep{Coutanche2011Multi} or multivariate statistical methods (e.g. multivariate pattern analysis) \citep{kloppel2012diagnostic, wolfers2015estimating}. These methods provide coefficients of variables showing the significance of biomarkers and enable the generation of statistical models with high diagnostic or predictive potential by focusing on patterns of brain changes that are distributed across multiple regions in disordered states. 
ML techniques have also been commonly used to select important features for brain disorder classification \citep{orru2012using, de2019machine}. A model that generalizes well to the test set and other unseen data points would be expected to have learnt optimal coefficients for each feature and feature importance could be inferred from these values. Features with the highest importance scores could represent potential biomarkers.

With the advent of deep learning techniques, multiple layers of non-linearities are introduced to learn complex relationships between the input and outputs. Furthermore, GNNs make it possible to model FC directly, obviating the need for feature engineering. Although they are often viewed as black boxes, the DNN's decisions could be analysed via model explainability algorithms such as gradients \citep{chen2022adversarial}, IG, or class activation mapping (CAM) \citep{qin2022using}.  

To consolidate the findings from these studies, one could conduct meta-analyses by calculating the contribution of identified biomarkers to specific disorders. For example, BrainMap \citep{fox2005brainmap} is a database of MNI coordinates for activation foci consolidated from thousands of experiments. Researchers can extract existing relevant studies from BrainMap using specific keywords, revealing experiments where both activated and non-activated ROIs were pinpointed and indicating their relevance to various disorders. Subsequently, methods such as the Naive Bayes classifier can be used to determine the probability of disorders associated with these ROIs \citep{yarkoni2011large}. 

Despite such progress in modelling techniques and efforts to curate larger datasets, biomarkers for brain disorders remain elusive. 
For instance, efforts to identify diagnostic biomarkers for depression could not arrive at any consistent depression biomarker despite extensive efforts \citep{winter2024systematic}. 
Thus, in the next sections, we introduce the literature on model explainability techniques (both model-agnostic and GNN-specific) with the goal of identifying research gaps (in existing fMRI studies on biomarker discovery) that could be addressed to improve the quality of these potential biomarkers.

% For example, SVM is a supervised learning model with associated learning algorithms (and kernels for learning linear or non-linear relationships) that allows for classification and regression analysis. SVMs have yielded high accuracies ($\sim90\%$) when applied to small fMRI datasets ($n < 100$) of SZ patients and highlighted ROIs with significantly lower brain activities \citep{liu2018abnormal}. Random forest (RF) exhibits significant advantages over other ML methods in terms of their capacity to handle highly non-linearly correlated data, simplicity of hyperparameter tuning, and robustness to noise \citep{sarica2017random, fredo2018diagnostic}. Their variable importance can be evaluated using a variety of methods, including calculation of the prediction power of selected features in classification using the impurity reduction principle.

\subsection{Properties of model explainability techniques}

Many model explainability algorithms have been proposed and they have been covered by numerous review papers \citep{linardatos2020explainable,tjoa2020survey,marcinkevivcs2023interpretable, yuan2022explainability}. We provide a brief summary of these feature attributors and place greater focus on key insights that are relevant to biomarker discovery. Before delving into the details of each attributor, we note several properties of attributors that are useful to characterise them.

Existing research on model explainability can be separated into three paradigms: (i) `glass box' (intrinsically interpretable), (ii) `black box' (reliant on post-hoc explainability methods), (iii) `grey box' (some interpretation possible, with careful design of the model) \citep{ali2023explainable}.
On one end of the spectrum, `shallow' models such as linear regression are intrinsically interpretable. In the case where all input features have the same scale, biomarkers can be extracted by identifying features that have the largest coefficients assigned to them by the model fitting process.
On the other end of the spectrum, deep learning models learn complex and non-linear relationships that cannot be easily interpreted. They often rely on post-hoc model explainability algorithms that are applied after model training. These algorithms typically generate scores based on some form of gradient computation (with respect to the input) or perturbation.
In between these two extremes, some complex models such as fuzzy rule-based systems and Bayesian networks can provide a limited extent of interpretability \citep{ali2023explainable}. The use of attention scores as well as graph pooling could also be grouped under this category. 

While `glass box' methods are desirable, most of these methods only capture linear relationships, which is likely to be of limited use for disease studies. Unlike how studies on using fMRI data to predict phenotypic information have shown that non-linearities do not give much improvement over linear models, deep learning models have shown better performance for disease classification and prediction of clinical test scores \citep{zhang2020survey}. 
However, many of these models fall under the `black box' category and more research is needed to create `grey box', or even `glass box' alternatives.

All three paradigms of attributors produce attribution scores that can be classified into two categories: local (`instance-level') and global (`model-level'). Local explanations are specific to the input data provided to the model (i.e. each sample has its own attribution scores), while global explanations apply broadly to the entire model (all samples share the same attribution scores). In the context of biomarker discovery, local explanations could be more desirable for clinical use if individual insights are found to be reliable. Additionally, global explanations are unlikely to be helpful for very heterogeneous diseases since heterogeneous diseases would not be fully described by a single set of global attribution scores.

\subsection{Model explainability for GNN}

\begin{figure*}
	\centering
		\includegraphics[width=\textwidth]{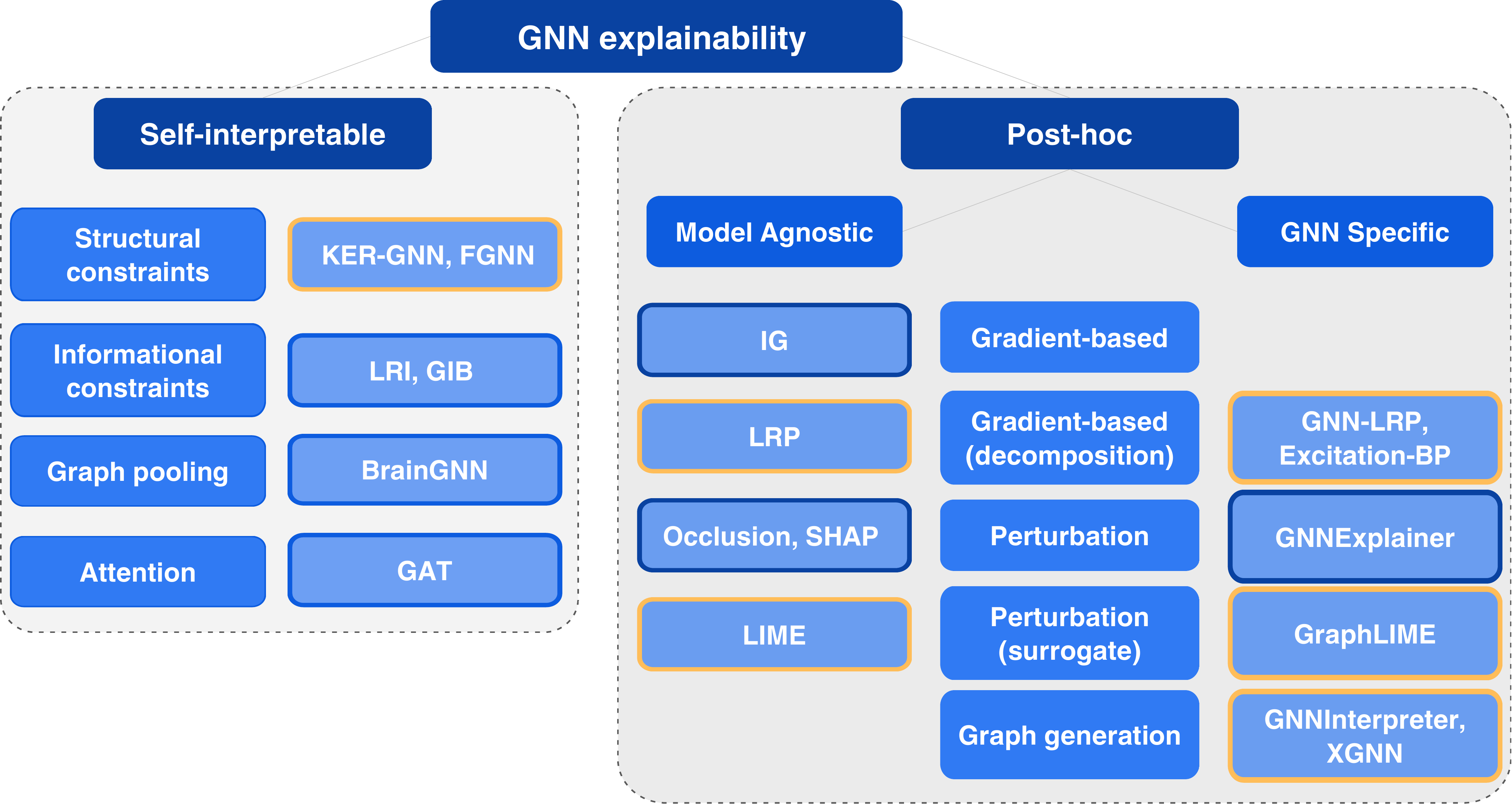}
	\caption{Taxonomy of attributors applicable to GNNs. Generally, post-hoc methods (typically `black box') have a larger range of algorithms than self-interpretable (typically `grey box') ones. Methods that have not been explored in fMRI studies are highlighted in yellow.}
	\label{fig:attributors}
\end{figure*}

GNNs can be difficult to interpret due to the non-Euclidean contextual information in the graph (node features and edge weights) that needs to be taken into account when computing attributions. 
For BG, attributions can be produced at the level of nodes (i.e. which ROIs contribute the most to the disorder), edges (i.e. which pairs of ROIs contributed most), and node features. 
For PG, attributions can be produced at such granularity too, but they carry a different meaning: nodes would correspond to patients, edges correspond to pairs of patients and node features would typically correspond to some form of imaging data.

Existing model explainability methods for GNNs can be split into two major groups: self-interpretable and post-hoc. 
Self-interpretable methods provide explanations simultaneously with the model predictions, while post-hoc methods are only applied after model training is complete. 
Self-interpretable methods typically include constraints to extract an informative subgraph, or architectural designs where weights that prioritise a subgraph are learnt. 
Post-hoc techniques can be further split into model-agnostic and GNN-specific techniques. Model agnostic algorithms can be applied to any deep learning model (and for GNNs, regardless of their internal structure), but some of them have been further extended to capture the graph structure \citep{pope2019explainability}.
GNN-specific methods explicitly consider the graph structure when generating the explanations. 
Model agnostic methods can be categorised into two major groups: gradient-based and perturbation-based.
GNN-specific methods follow such a characterisation too but also have unique ones such as techniques based on graph generation. 
Figure \ref{fig:attributors} summarises the taxonomy of GNN explainability methods and in the following subsections, each subcategory will be explained in detail and discussed in the context of biomarker discovery.

\subsubsection{Self-interpretable - Structural constraints}

Drawing inspiration from CNNs, kernel GNN (\textbf{KerGNN})  introduces learnable graph filters (akin to convolutional filters in CNNs) into the messaging passing process in GNNs \citep{feng2022kergnns}.
Contrary to the rooted subtree approach used in GNNs that are based on the message-passing paradigm, KerGNN updates each node's embedding based on subgraphs that are centered on the node. 
This is done via the use of graph kernels (specifically, the random walk kernel) to measure the similarity between subgraphs and the graph filters. Not only does such an approach make GNNs more expressive (going beyond the 1-dimensional Weisfeiler-Leman (1-WL) limit), it also provides additional interpretability via the learnt graph filters, just like how convolutional filters can be visualised. 

Such an approach would be useful in the case of disorders that are poorly understood as novel insights could be drawn from these visualisations. However, if there is some existing biological knowledge available, Factor GNN (\textbf{FGNN}) \citep{ma2019incorporating} could be considered. 
Unlike usual deep learning models where the hidden nodes in deep learning models do not have a physical meaning, nodes in FGNN represent biological units that could be either observable variables or latent variables. Through such representations, FGNN directly incorporates biological knowledge as the inductive bias into the model.
 
While there have not been any applications of FGNN on fMRI datasets (the original paper applied it on genetic data), one example of how it can be applied is to use functional brain modules (i.e. a group of ROIs that are typically well-connected to each other, yet less connected to ROIs not in the module) as the latent variables and ROIs as the observable variables. This forms a factor graph that can be used to guide the construction of the neural network. Unlike a fully connected layer, input nodes (observable variables) representing ROIs will only be connected to the intermediate layer (latent variables) if the ROI belongs to the module. Then, the hidden layer could be used as input to a fully connected layer to make predictions (e.g. clinical outcomes), or be passed to another stack of factor graphs, forming a deep network.

\subsubsection{Self-interpretable - Informational constraints}

Instead of introducing architectural designs that guide the learning process (and identify salient subgraphs), methods based on information constraints introduce information bottlenecks such that the mutual information (MI) between the labels and the discovered subgraph is maximised, while keeping the MI between the original graph and subgraph below a predefined threshold. While the former can be approximated via cross-entropy loss, the latter is estimated via techniques such as learnable randomness injection (\textbf{LRI}) \citep{miao2022interpretable} and graph information bottleneck (\textbf{GIB}) \citep{yu2021recognizing}. The latter has been further developed in the fMRI biomarker discovery literature by BrainIB \citep{zheng2024brainib}. It extends GIB by considering the effects of edges (not just nodes) during subgraph discovery.

\subsubsection{Self-interpretable - Graph pooling}
\label{sec-pooling}

Graph pooling is often performed in GNN architectures, especially in graph classification tasks where node features have to be condensed to a lower dimensionality or a single vector. 
Pooling techniques can be grouped into two categories: (i) flat pooling, where a graph-level representation is generated in one step, and (ii) hierarchical pooling, which gradually coarsens the graph by clustering nodes together or dropping some of them \citep{liu2022graph}.

Several pooling techniques customised for FC data have been proposed. 
\cite{li2021braingnn} proposed a node / ROI pooling layer (R-pool) in their \textbf{BrainGNN} architecture. R-pool projects the node feature embeddings to a learnable weight vector and retains nodes with the highest scores. Hierarchical pooling approaches that consider functional modules have also been proposed \citep{mei2022modular}. In this work, three levels of hierarchy were used: (i) ROIs belonging to the same sub-network (e.g. Yeo 7-network parcellation \citep{yeo2011organization}) and brain hemisphere, (ii) the pair of matching sub-networks from each hemisphere, (iii) combining all sub-networks into a whole brain network. Weights from the final pooling layer were used to identify the sub-networks that contributed most to the model's decision.

\subsubsection{Self-interpretable - Attention}

The widespread use of attention for model interpretability has also been present in the GNN literature, most popularly via self-attention in \textbf{GAT}. Attention scores have been used to identify salient FC features as well \citep{zhang2022identifying,yu2022graph}.
However, there has been much debate in natural language processing (NLP) research about whether attention scores provide meaningful explanations \citep{jain2019attention,wiegreffe2019attention,serrano2019attention,bai2021attentions}. 
In NLP applications, attention scores were found to not correlate well with multiple gradient-based approaches on recurrent neural networks and different attention distributions can lead to equivalent predictions \citep{jain2019attention}. However, Wiegreffe \emph{et al.} \citep{wiegreffe2019attention} argue that explanations should be further categorised into `plausible' or `faithful' explanations. Their results provide additional support that attention does not provide faithful explanations but does not invalidate claims that attention provides plausible explanations. These results suggest that greater care should be taken when using attention scores to discover potential biomarkers. This has been further substantiated in a recent mathematical study of a single-layer multi-head transformer architecture, which showed that post-hoc methods provide more insightful explanations than attention weights \citep{lopardo2024attention}. Further research is needed to examine the validity of attention scores for biomarker discovery applications. 
For instance, Safai et al. \citep{safai2022multimodal} found that the attention weights of a GAT model used for Parkinson's disease classification could be used to identify which brain regions contribute to the classification accuracy, but they vary significantly across attention heads. While they analysed each head separately, it is currently still unclear how findings from different attention heads of the same GAT model should be reconciled.

\subsubsection{Post-hoc - Gradient-based}
A large variety of gradient-based approaches have been proposed. Integrated Gradients (\textbf{IG}) \citep{sundararajan2017axiomatic} will be discussed here due to its versatility (works on most deep learning models where gradients can be calculated) and widespread use. 
IG was developed to address the issue of saturation in gradient-based attribution methods. With saturation, the output is no longer sensitive to small changes in the input features, making it difficult to interpret which features are responsible for the prediction of the correct class \citep{sturmfels2020visualizing}. IG avoids this issue by accumulating gradients from interpolated points (controlled by $\alpha$) between a baseline ($x'$) and the input data ($x$). In the context of disorder classification, the baseline can be the average data of all healthy subjects. For a given feature $k$, IG is defined as:

$$IG_k (x) = (x_k - x'_{_k}) \times \int_{\alpha=0}^{1} \frac{\partial f(x'_{_k} + \alpha \times (x - x'_{_k}))}{\partial x_k} d\alpha.$$

Attribution scores from IG provide a local measure of how much the feature contributed to the model's prediction and could be potentially useful for producing personalised insights. At present, its use is still limited to group-level insights, e.g. identifying site-specific and site-invariant biomarkers of SZ \citep{chan2022semi}.
This is in part due to how IG can also produce noisy pixel attributions in features unrelated to the predicted class. 
Modified versions of IG, such as GuidedIG \citep{kapishnikov2021guided}, 
have been proposed to address these issues. 
Features with high gradient scores have a greater impact on predicting the class of the model than features with low gradient scores. 
Thus, GuidedIG reduces noise in the attribution results by using an adaptive path technique that only incorporates a subset of features with high gradient scores.

\subsubsection{Post-hoc - Decomposition}

Decomposition-based approaches have some overlaps with gradient-based approaches but differ in the way the scores are computed. Instead of computing the gradients directly with respect to inputs, scores are decomposed starting from the output layer and propagated backwards in a layerwise manner based on pre-defined rules. 
Layerwise relevance propagation (\textbf{LRP}) is one such example. Many forms of LRP exist and the variant called $\epsilon$-LRP will be discussed \citep{montavon2019layer}. 
Starting from the output layer, a score $s$ is assigned to a neuron based on the logit, i.e. 
$$s^{l}_{i} = \frac{h_{ji}}{\sum_{i} h_{ji} +  \epsilon (\sum_{i} h_{ji})} s^{l+1}_{i},$$
where $h_{ji}$ refers to the output from neuron $i$ in layer $l$ to neuron $j$ in layer $l+1$. This is computed layerwise, reallocating the prediction score until the input layer is reached. The total relevance score of $s$ is always preserved for each layer. 

LRP does not consider the adjacency matrix in its computations. To address this, \textbf{GNN-LRP} \citep{schnake2021higher} distributes scores to different graph walks (and thus have higher computational complexity). \textbf{Excitation-BP} is very similar to LRP, but it views the decomposition process from a probability standpoint. Overall, recent decomposition-based approaches like GNN-LRP have not been well-studied in fMRI, with usage mainly limited to LRP. \citep{yan2017discriminating}

\subsubsection{Post-hoc - Perturbation}

Perturbation-based approaches introduce changes to the input with the motivation that if important features are still retained, the outputs should remain similar. In its simplest implementation (\textbf{Occlusion}), this involves masking features one by one and the feature that results in the largest change in output would be deemed to be the most important. \textbf{SHAP} \citep{lundberg2017unified} takes this idea to completion by considering all feature subsets (i.e. $2^k$ combinations), so as to compute Shapely scores that have been proven to be the unique solution that fulfills the criteria of local accuracy (model training on best feature subset should have similar predictions with the original model), missingness (features not in the best subset should have no impact on the model output) and completeness (attribution score should not decrease when a different model, where the feature contribution does not decrease, is used). Computing this is too computationally expensive, thus it is achieved via approximation techniques. 

In the context of graphs, such perturbations can be performed by discovering subgraphs. 
\textbf{GNNExplainer} \citep{ying2019gnnexplainer} produces local explanations for GNN predictions by selecting a small subgraph from a given input graph and identifying important node features. 
Subgraphs are generated by randomly masking nodes in the graph and observing the resulting changes in the model's prediction. A soft mask (i.e. continuous values, not binarised) containing learnable weights is used.
Important node features (that are in the nodes within the subgraph) are identified by a binary feature selector. 
The mask and feature selector are optimised by maximising the mutual information between the original model predictions and the model's predictions given the masked graph. 
One limitation of GNNExplainer is that the subgraph must be connected, which might not always be applicable to disease biomarkers. Nevertheless, it has been quite popularly used in fMRI studies \citep{gallo2023functional,hu2021gat}.

\subsubsection{Post-hoc - Surrogate}

Surrogate-based approaches have some overlaps with perturbation-based approaches as they tend to rely on perturbations too. However, a distinctive feature is the use of simpler and interpretable models (often linear) to approximate the original complex model. 
This is possible as it limits the approximation to a local neighbourhood and analyse the model predictions of perturbed inputs within this neighbourhood. For instance, local interpretable model-agnostic explanations (\textbf{LIME}) \citep{ribeiro2016should} trains a surrogate model on the dataset of perturbed points, weighing them based on their proximity to the chosen data point. 

\textbf{GraphLIME} \citep{huang2022graphlime} is a non-linear version of LIME, a key difference being that it uses Hilbert-Schmidt Independence Criterion (HSIC) lasso, a kernel-based non-linear interpretable feature selection algorithm. 
The algorithm first computes the importance of each feature in each node by considering the features of the target node and features in the N-hop neighbouring nodes. The given target node will aggregate the information from N-hop network neighbours to identify the most significant features. The HSIC lasso method is used to train a linear interpretable model to represent the relationship between the features and target node prediction. Subsequently, the coefficients from the linear interpretable model will be used to identify the top few features that are important for model prediction based on the coefficients. Thus far, no research on FC has used such techniques for biomarker discovery.

\subsubsection{Post-hoc - Graph generation}

Attributors based on graph generation bear some similarities with GNNExplainer as both identify salient subgraphs. However, generation-based approaches arrive at the subgraph via generative approaches, instead of perturbation.
\textbf{XGNN} \citep{yuan2020xgnn} interprets GNNs using a graph generator to identify important graph subgraphs. The generator is trained using reinforcement learning (RL) and validity rules are defined by pre-existing knowledge, making them less suitable for biomarker discovery. On the other hand, \textbf{GNNinterpreter} \citep{wang2023gnninterpreter} do not require pre-existing knowledge as it optimises the choice of subgraph by maximising the similarity between embeddings from the important subgraph with that of the average graph embeddings in the target class. Both attributors are different from all other attributors discussed above as they produce global explanations (i.e. one set of explanations for the whole model, across all data points). Thus far, no studies on FC have used these post-hoc attributors based on graph generation for biomarker discovery.

\subsubsection{Summary}

\begin{table}[]
\caption{Key characteristics of attributors. Methods above the line are self-interpretable, while those below are post-hoc methods. 
}
\begin{tabular}{lllll}
\hline
Attributor & Approach & Granularity & Target & Trainable\\
\hline
KER-GNN & Structural & Instance & Node features & Yes\\
FGNN & Structural & Instance & Node features & Yes\\
LRI & Informational & Instance & Subgraph & Yes\\
GIB & Informational & Instance & Subgraph & Yes\\
BrainGNN & Pooling & Instance & Node & Yes\\
GAT & Attention & Instance & Edge & Yes\\
\hline
IG & Gradients & Instance & Node features & No\\
LRP & Decomposition & Instance & Node features & No\\
GNN-LRP & Decomposition & Instance & Edge & No\\
Excitation BP & Perturbation & Instance & Node features & No\\
Occlusion & Perturbation & Instance & Node features & No\\
SHAP & Perturbation & Instance & Node features & No\\
GNNExplainer & Perturbation & Instance & Subgraph, Node features & Yes\\
LIME & Surrogate & Instance & Node features & Yes\\
GraphLIME & Surrogate & Instance & Node features & Yes\\
GNNInterpreter & Graph generation & Model & Subgraph & Yes\\
XGNN & Graph generation & Model & Subgraph & Yes\\
\hline
\label{tab:attri}
\end{tabular}
\end{table}

Figure S\ref{fig:pie_attri} provides a visual breakdown of the attributors used in the papers included in this review. It is evident that attributors based on informational constraints, pooling, attention, gradients, and perturbation have been used on FC datasets.
Notably, several attributors that are not graph-based have also been tested, e.g. trainable masks, clustering, weights, and variants of Gradient-weighted Class Activation Mapping. 
However, there remains much room to explore alternative attributors based on graph generation, decomposition, surrogates, and structural constraints. 

When choosing an attributor for biomarker discovery, it is crucial to first understand which types of explanations the method can provide (what granularity, what attribution targets are possible, and whether the attributor requires training). 
As a useful reference for future research, Table \ref{tab:attri} summarises these key characteristics of attributors highlighted in Figure \ref{fig:attributors}. Notably, GNN-specific attributors (e.g. GNNExplainer) have the advantage of being able to identify subgraphs, while generic attributors (e.g. IG) are typically limited to node features.

Overall, there are still several types of attributors not explored in fMRI studies yet. Furthermore, thorough benchmarking studies would be needed to verify the robustness of these attributions, beyond the existing predominant practice of qualitatively comparing the top-k attributions against existing work.

\subsection{Evaluation of strengths and weaknesses}

Self-interpretable attributors have a clear advantage over post-hoc methods as they do not need a separate attributor, which at times requires training yet another deep learning model. 
However, many of them fundamentally rely on the model learning some form of weight matrices, which are designed to be consistent with the input feature dimensions, and thereby seen as a measure of importance. 
This is sensible for linear models, e.g. the coefficients learnt by a linear regression model with features of the same scale could be interpreted as a measure of feature importance. 
But when this is applied in the context where non-linear functions are used to transform both the input and outputs of the layer where the weights are extracted from, it might be too simplistic as these representations are not humanly understandable at present, even though they might still seem to correspond to individual ROIs. Given the non-linear transformations, there is little guarantee that a node with higher weight would contribute more greatly to the model's final decision.
While existing studies attempt to demonstrate the correctness of their attributions cross-referencing with other research studies, the uncertainties brought about by the lack of understanding of these opaque operations warrant stricter evaluation techniques that go beyond the current practice.

Creating a foolproof technique to evaluate attributors could be challenging due to the lack of ground truth as well as how many of the novel GNNs introduce modifications that are specific to the human brain. The use of other baseline graph datasets with known ground truth, such as MUTAG, could be considered as a proxy for methods that are not tied to neurosciences. In the other scenario, one could attempt to demonstrate the validity of their techniques via synthetic datasets \citep{agarwal2023evaluating}. 
Post-hoc methods could also be applied on top of these self-interpretable methods to assess the extent of agreement between both methods. 

Comparing model agnostic and GNN-specific post-hoc explainability techniques, the most critical factor is often the target that the attributor is able to compute contributions for. In that regard, GNNExplainer provides the most flexibility as it is able to compute both motif-level and node-level contributions. On the other hand, many methods are only limited to a single target type. While several studies have demonstrated how contributions at the level of node features (i.e. edges, when connection profile is used) can be consolidated to nodes, the appropriateness of these aggregations should be studied in a more robust manner, e.g. testing the validity of such operations on synthetic data with known ground truth.

\section{Evaluation of attributions}

Better biomarker discovery tools are needed as few potential biomarkers turn out to be effective in clinical settings \citep{parkes2020towards}. To bring us closer to this goal, one solution could involve developing objective means of assessing the robustness of attribution scores produced by these attributors. Robustness entails the expectation that (i) attributions should capture valid signals (e.g. features that are indeed idiosyncratic characteristics of the disorder) and that (ii) the scores should be reasonably invariant to non-disorder related factors while retaining sensitivity to disorder-related changes. 
Verification of validity is a challenge for psychiatric disorders as the ground truth is often unavailable. However, it is possible to assess the latter with appropriate metrics defined in the context of deep learning models. For instance, it would be unreasonable for attributions to vary significantly just by changing the size of the hidden channels of a GNN - salient features highlighted by such models would be deemed to be unreliable.
The complexity of deep learning models necessitates a comprehensive framework to characterise these factors of variation.

\subsection{Existing evaluation techniques}

\begin{figure*}
	\centering
		\includegraphics[width=0.8\textwidth]{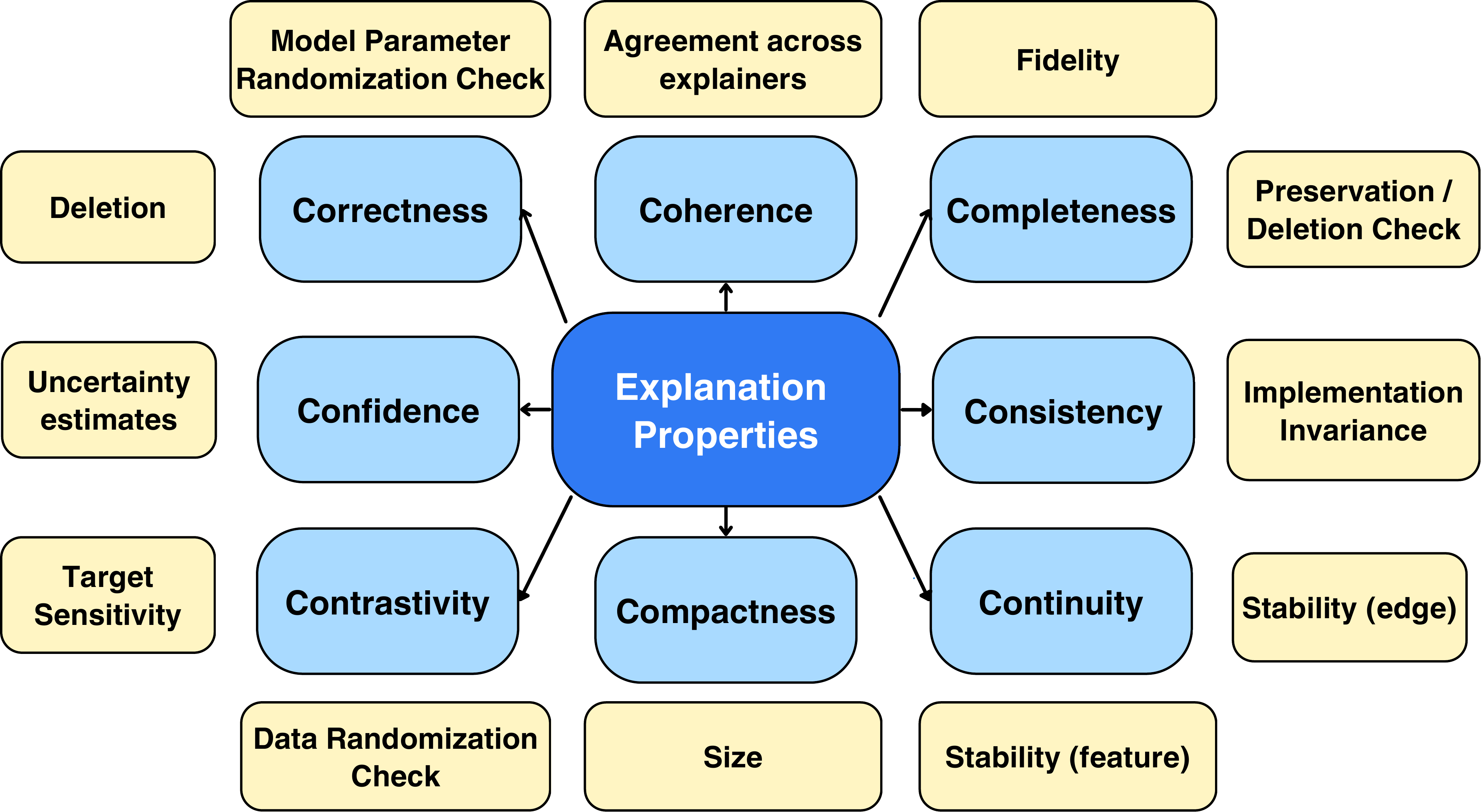}
	\caption{A subset of the Co-12 properties \citep{nauta2023anecdotal} used to evaluate feature attribution scores produced by attributors, deemed to be relevant for biomarker discovery from fMRI datasets.}
	\label{fig:evaluators}
\end{figure*}

Several methods of evaluating generic attributors \citep{nauta2023anecdotal} as well as GNN-specific attributors \citep{agarwal2023evaluating,li2022survey} have been proposed. Nauta \emph{et al.} \citep{nauta2023anecdotal} proposed a framework named `Co-12' which encompasses 12 desired properties of attributors (such as Correctness, Completeness, Consistency, etc.). In this section, we discuss properties that are relevant to biomarker discovery and extend the analysis to GNN-specific attributors. 
Figure \ref{fig:evaluators} illustrates the 8 chosen properties and metrics that can be used to measure a model's performance with respect to the corresponding property. An explanation of why the 4 other components are not included can be found in the Appendix. 
Additionally, we note that not all 8 properties have been covered by existing fMRI studies. For those that are covered, the relevant references have been added to the details presented in each subheading below.

Each property below encompasses one or more evaluation metrics. Many metrics involve comparing two distributions (e.g. original attributions versus attributions produced in a different setting) and measuring the distance between them. This can be computed via Hellinger distance, which has an easily interpretable range (0 = perfectly similar, 1 = completely different distributions).

\subsubsection{Correctness} Sanity checks on the attributors can be performed via \textbf{model parameter randomisation check}, which verifies whether the explanation of a trained model is different from a randomly initialised untrained model. 
Other forms of evaluating correctness involve \textbf{deletion}: changes in model outputs are computed for each feature subset (in the simplest case, independently removing each feature in the node vector) and then correlated with the importance scores. However, this could be too tedious for deep learning models applied on connectome datasets especially if many ROIs are used. 

\subsubsection{Completeness} Salient subgraphs or feature subsets identified by the attributors should not lose too much information as compared to the original input. This can be assessed via \textbf{preservation check} (whether using the selected features as input results in the same model prediction) and \textbf{deletion check} (not using the important features results in a different prediction). 

\textbf{Fidelity} is based on a similar idea but quantifies the difference by measuring the change in probabilities (as opposed to a simple change in prediction of classes). Note that this can be done both at the level of node features and at the level of subgraphs (in the case of graph datasets like FC). In recent fMRI studies, this has been analysed at the level of node features \citep{hu2021gat,safai2022multimodal,menon2023asdexplainer}.
An example of how Fidelity+ (represented by $v_{F+}$) can be computed is shown below:

\begin{equation}
    v_{F+} = \frac{1}{N} \sum_i (f(X_i) - f(X_i^{+})), 
\end{equation}

where $f(\cdot)$ is the model under study, $X_i$ represents the node features of subject $i$ from a dataset of $N$ subjects whereas $X_i^+$ represents the subset of node features where features with the highest attribution scores are removed.

\subsubsection{Consistency} Explanations should be robust across most variations in model implementations. While deep learning models can involve many hyperparameters that have profound and poorly understood changes to the model, a basic form of assessment for \textbf{implementation invariance} could entail checking whether two models with the same architecture, but having different initialisation (i.e. randomly initialised weights), will produce similar outputs and explanations.

\subsubsection{Continuity} Given a small change in the input that still produces a similar model output, it would be reasonable to expect that the explanation will remain similar too. \textbf{Stability} ensures this by introducing random noise to the input features. For graphs, additional changes can be introduced by randomly switching edges while keeping the number of edges constant. Both explanations (before and after perturbation) can then be compared to determine if the scores have good continuity. 

\subsubsection{Contrastivity} Explanations produced for data samples belonging to different classes would be expected to differ. This is verified via a \textbf{target sensitivity} check, where the mean explanations for each class can be computed and then compared across classes. Alternatively, a \textbf{data randomisation check} can be performed by randomising the labels and verifying that the explanations are changed. Given that biomarker discovery techniques based on ML techniques are often based on disorder classification tasks, it would be important to check if the scores produced have high contrastivity.

\subsubsection{Compactness} Explanations should not overwhelm the user and this is especially relevant for FC datasets, where hundreds of ROIs and thousands of connections are analysed simultaneously in multivariate studies. 
For instance, while severe diseases might affect widespread areas of the brain, having an explanation that gives similar scores to an overwhelming number of ROIs or edges might not be as useful as explanations that correctly emphasise a small group of features. Computing the \textbf{size} of explanations is straightforward for attributors that produce subgraphs (e.g. number of connections divided by the number of edges in a complete graph), but would require thresholding in the case of most gradient-based approaches (since a score is given to each feature, but some scores could be very close to 0). This has been examined in \citep{menon2023asdexplainer}, where SubgraphX was shown to have higher sparsity than GNNExplainer and PGExplainer. 

\subsubsection{Confidence} While almost all attributors discussed above do not produce uncertainty estimates and many of their scores have no natural variations (e.g. Saliency is based on gradients with respect to inputs, which would be fixed given the same model and input), measures of confidence could be computed based on the logits of the predictor. One example of an implementation is proposed by Atanasova \emph{et al.} \citep{atanasova2024diagnostic}, where class probabilities from the predictor are compared against a predicted confidence value estimated by fitting a logistic regression model to saliency distance values (which computes the distance between saliency across in each class). A similar approach could be used for biomarker discovery applications.

\subsubsection{Coherence} Since the ground truth is often unknown in biomarker discovery, coherence would be focused on the extent of agreement between attributors. Biomarkers that are present across attributors based on different approaches could be deemed to be more robust. However, further studies need to be conducted to determine the attributors to use (e.g. attributors with low stability should not be considered). Many existing fMRI studies that analysed robustness chose to evaluate coherence, revealing that attribution can vary significantly \citep{li2023classification} but there are still instances where there are ample overlaps \citep{gallo2023functional}.

\subsection{Summary}

Overall, we found that only 6 out of the 65 fMRI studies covered in this review have used objective metrics to evaluate the robustness of the potential biomarkers highlighted by their model. This is in line with past reviews on the use of interpretable deep learning techniques in more general medical settings which found that only a minority of existing studies (13 out of 75) evaluated robustness \citep{munroe2024applications}. 

Linking back to the second question mentioned in the introduction (how existing studies evaluate the robustness of their biomarkers), we found that studies that assessed the robustness of attributors were largely limited to aspects such as coherence, completeness, and compactness. 
Thus, there are still many other aspects such as continuity, contrastivity, and correctness that should be considered in future studies. An example of how these can be studied is demonstrated in \citep{girishrobustness}. In essence, metrics that can be customised to focus on top-k features (e.g. Fidelity) can be used to measure the robustness of attributions, while other more generic metrics could serve as a sanity check of the model.

\section{Application of GNNs on disorder prediction and biomarker discovery from fMRI data}
\label{sec:survey}

Biomarkers are biological signatures that can be objectively measured and used to indicate physiological processes, pharmacological responses, or disease status \citep{jain2010handbook}. Diagnostic biomarkers refer to a specific type of biomarker that can be used to identify patients with a particular disorder or condition \citep{califf2018biomarker}.
In this paper, we focus our discussion on potential diagnostic biomarkers of brain disorders derived from fMRI via GNNs. 

The salient features detected from trained models are potentially disease biomarkers, but it is uncertain whether they capture true signals of the disorder, or are mere artefacts of noise. 
To ascertain the usefulness of these potential biomarkers, they need to be evaluated against the following traits: easily accessible, reproducible, specific, and sensitive. Most importantly, it should only change depending on the state of the disease and be unchanged when unrelated factors are varied \citep{aronson2017biomarkers}. Most existing studies on biomarkers for brain disorders have not attained such high standards yet and bridging this gap has been a priority for neuroimaging research in recent years \citep{parkes2020towards}. 

In the following subsections, we summarise key biomarkers identified from the studies included in this review. A key goal of this review is to identify salient features that are reproduced across multiple studies. In view of this, we focus on papers that include resting-state fMRI due to its widespread availability as compared to task fMRI scans. 
Taking into account that several papers tested their models on multiple datasets and disorders, we found that ASD dominates existing research (53.8\% of papers considered in this review) along with MDD (26.9\%), while SZ (14.1\%) and ADHD (5.1\%) are relatively rare.

Salient features present in multiple studies would be more promising biomarkers of the disorder as they exhibit greater robustness than non-reproducible ones. 
However, this is still a challenging task to identify them as they can vary in granularity: ROI, connections, module, modular connections, and temporal features. 
Furthermore, the choice of brain atlas and attributor are often different. 
Thus, to aid readability, the following tables are provided in the Appendix: the names of brain atlases used are abbreviated to make it easy to note how many ROIs are involved and the full names can be found in Table \ref{tab:atlas}. Also, Table \ref{tab:abbrev} provides a mapping of the abbreviations to the full names,
while full names for abbreviations of functional modules and ROIs can be found in Table \ref{tab:mod} and Table \ref{tab:roi} respectively.

\subsection{Attention Deficit Hyperactivity Disorder}
\label{sec-adhd}

\begin{table}[ht]
\caption{Summary of findings from ADHD studies that identified potential biomarkers. `Size' refers to the size of the dataset. When modalities beyond sFC are used, they are marked with [d] (dFC) or [M] (multimodal). `Type' refers to the granularity of attributions.
}
\begin{tabularx}{\textwidth}{L{0.5}L{0.70}L{0.5}L{0.80}L{2.5}}
\toprule
\textbf{Reference} & \textbf{Dataset (Size)} & \textbf{Atlas} & \textbf{Attributor (Type)} & \textbf{Salient features identified} \\
\toprule

\multicolumn{5}{c}{\textbf{ADHD}} \\
\midrule

\cite{yu2022graph} & CNI-TLC (240) & CC200 & Attention (ROI, connection) & 
\textbf{Regions} (-): FP, IFG, MFG, MTG, posterior central gyrus, SOG;
\textbf{Regions} (+): Brainstem, FOG, precuneus, putamen, TP;
\textbf{Connections} (-): IFG-MTG, MFG-FP, MFG-posterior central gyrus, posterior central gyrus-SFG;
\textbf{Connections} (+): Cerebellum-fusiform gyrus, FOG-Precuneus, PARAH-STG
\\ [7.5mm]

\cite{zhao2022dynamic} & ADHD-200 (603) & AAL116 & Weights (ROI, connection) & 
\textbf{Regions}: Frontal lobe, occipital lobe, subcortical, temporal lobe, posterior-fossa (cerebellum);
\textbf{Connections} (-): right rolandic operculum - right Heschl gyrus;
\textbf{Connections} (+): left precuneus - right cerebellum, left STG/TP - right medial SFG
\\[7.5mm]

\cite{luo2024knowledge} & ADHD-200 (506) & AAL116 & Attention (ROI) & 
\textbf{Regions}: bilateral olfactory cortex, bilateral MFG (orbital part), left gyrus rectus, bilateral posterior cingulum, bilateral putamen
\\[7.5mm]

\cite{zhang2023gcl} & ADHD-200 (506) & Multiple & Trainable mask (ROI) & 
\textbf{Regions}: right PrCG, right thalamus, and right MFG
\\[7.5mm]

\bottomrule

\end{tabularx}
\label{tab:adhd}
\end{table}

ADHD is a syndrome characterised by the presence of inattentive and/or hyperactive (and impulsive) behavior to an extent that is age-inappropriate and often affects social, academic, and occupational performance. It is widely considered to have 3 broad subtypes based on behaviour (combined, inattentive, and hyperactive/impulsive).
It typically commences from early childhood years but could persist into adulthood.

Relative to other complex psychiatric disorders, the neural underpinnings of ADHD have been more clearly elucidated: disruption to the response inhibition (dorsomedial frontal cortex, anterior insula/inferior frontal cortex) \citep{aron2004inhibition} and neural reward processing (ventral striatum) circuits are associated with ADHD \citep{plichta2014ventral}. Specifically, there is dysregulation of dopaminergic and noradrenergic systems in these regions \citep{del2011roles}. Even though there are treatments like psychostimulants and neuromodulation, the cause of ADHD is still poorly understood and diagnosis is complicated by the overlap of symptoms with other related conditions. Thus, it is worth exploring whether fMRI studies could identify regions and functional connections that are implicated in ADHD, beyond our current understanding of the disorder.

In this review, 4 studies on ADHD satisfied our inclusion criteria, as shown in Table \ref{tab:adhd}. They relied on datasets from CUNMET (Spain), ADHD-200 (a mix of sites from the USA and China), and CNI-TLC (with some overlaps with site KKI in ADHD-200). 
CUNMET contains a mix of combined and inattentive subtypes, ADHD-200 has a mix of all three subtypes and KKI has all three but is dominated by the combined subtype. Hyperactivity/impulsive is under-represented across all datasets. 
Overall, model performance is moderately high (mean accuracy of 71.8\% across the studies). Considering the moderate size of the datasets (mean: 374) used, these results suggest that fMRI has a moderate ability to discern between typical controls and ADHD patients. This remains the case even when including studies that did not perform biomarker analysis.

However, the analysis of salient features was less encouraging. While both Yu et al. and Zhao et al. (trained on different datasets) agree that ADHD subjects have weaker connections between the frontal lobe and temporal lobe, there is no agreement on the level of connections between ROIs. 
For instance, \cite{yu2022graph} highlighted MTG-IFG as a weakened connection, while \cite{zhao2022dynamic} highlighted the left temporal pole of STG - right SFG (medial).
Nevertheless, a majority of studies do implicate the middle frontal gyrus, which could be related to the response inhibition circuit. However, more studies need to be conducted before coming to any conclusion.

Additionally, since ADHD is heterogeneous, it would be more insightful for future studies to analyse salient features separately for each subtype. The presence of heterogeneity in the disorder suggests that biomarker reproducibility might not always be possible (e.g. different datasets are dominated by different subtypes of ADHD) and a subtype-level analysis would have more potential of highlighting replicable biomarkers.

\subsection{Autism Spectrum Disorder}
\label{sec-asd}

\begin{table}[ht]
\caption{Summary of findings from ASD studies that identified potential biomarkers. `Size' refers to the size of the dataset. When modalities beyond sFC are used, they are marked with [d] (dFC) or [M] (multimodal). `Type' refers to the granularity of attributions.
}
\begin{tabularx}{\textwidth}{L{0.5}L{0.70}L{0.5}L{0.80}L{2.5}}
\toprule
\textbf{Reference} & \textbf{Dataset (Size)} & \textbf{Atlas} & \textbf{Attributor (Type)} & \textbf{Salient features} \\
\toprule
\multicolumn{5}{c}{\textbf{ASD}} \\
\midrule

\cite{li2022te} & ABIDE (866) & MODL128 & Finetuning (ROI) & 
\textbf{Regions}: STG ; Cerebellum IV and V, central parieto-occipital sulcus, left superior temporal sulcus, left anterior intraparietal sulcus, cerebrospinal fluid (between superior part of SFG and skull)
\\[3.5mm]

\cite{zhang2022classification} & ABIDE (871) & HO112 & Attention (ROI) & 
\textbf{Regions}: IFG, PrCG, frontal orbital cortex, PARAH. 
\\[3.5mm]

\cite{wang2022mage} & ABIDE (949) & Multiple & Unclear (ROI) &
\textbf{Regions}: angular gyrus, PrCG, precuneus and thalamus 
\\[3.5mm]

\cite{zhu2021triple} & ABIDE (1112) & AAL116 & Pooling (ROI) & 
\textbf{Regions}: lateral PFC, lateral dorsal PFC, SPL 
\\[3.5mm]

\cite{shao2021classification} & ABIDE (871) & HO111 & Feature selection (connection) & 
\textbf{Connections}: evenly distributed across the brain, lower in ASD for 25/30 FC e.g. right cuneal cortex and right parietal operculum cortex \\[3.5mm]

\cite{wang2021graph} & ABIDE (1057) & CC200 & Occlusion (module) & 
\textbf{Modules}: DMN, FPN, VAN \\[3.5mm]

\cite{li2022te} & ABIDE (871) & Multiple & Clustering (module, modular connection) & \textbf{Modules}: CEN, DMN, SN ; \textbf{Modular connections}: CEN-SMN, CEN-SN, DMN-SMN, DMN-CEN, DMN-Visual \\[3.5mm]

% dfc

\cite{zhu2022contrastive} & ABIDE (613) [d] & HO110 & Feature selection (ROI) & 
\textbf{Regions}: lingual gyrus, MFG, SFG 
\\[3.5mm]

\cite{cui2023dynamic} & ABIDE (1035) [d] & CC200 & Gradients (ROI) & 
\textbf{Regions}: FP, precuneus, brain stem, paracingulate gyrus, lingual, OP, lateral occipital cortex, frontal orbital cortex \\[3.5mm]

\cite{chen2022invertible} & ABIDE (867) [d] & HO110 & Feature selection (connection) & 
\textbf{Connections}: right pallidum-right IFG, left frontal orbital cortex-left central opercular cortex, left supramarginal gyrus - right ITG \\[3.5mm]

\bottomrule
\end{tabularx}
\label{tab:asd}
\end{table}

\begin{table}[ht]
\caption{Summary of findings from ASD studies that identified potential biomarkers. `Size' refers to the size of the dataset. When modalities beyond sFC are used, they are marked with [d] (dFC) or [M] (multimodal). `Type' refers to the granularity of attributions.
}
\begin{tabularx}{\textwidth}{L{0.5}L{0.70}L{0.5}L{0.80}L{2.5}}
\toprule
\textbf{Reference} & \textbf{Dataset (Size)} & \textbf{Atlas} & \textbf{Attributor (Type)} & \textbf{Salient features} \\
\toprule
\multicolumn{5}{c}{\textbf{ASD (continued)}} \\
\midrule

%%%%%
\cite{chen2021attention} & ABIDE (1007) [M] & AAL116 & Gradients (ROI, connection) & 
\textbf{Regions}: (higher T1w): DMN, reward, memory and motor ; (higher ALFF) reward and motor ; 
\textbf{Connections}: inter \textgreater intra, low homotopic interhemispheric connection in limbic regions \\[3.5mm]

\cite{chen2022adversarial} & ABIDE (1007) [M] & AAL116 & Gradients (connection) & 
\textbf{Connections} (-): right MTG and multiple ROIs in the frontal, parietal, and occipital lobes; 
mix of higher and lower FCs between ROIs in the limbic regions to multiple other regions \\
%%%%% from above

\cite{li2020graph} & Biopoint (118) & DX148 & Clustering (ROI) & 
\textbf{Regions}: PFC, cingulate cortex
\\[3.5mm]

\cite{li2020pooling} & Biopoint (118) & DK84 & Pooling (ROI) & 
\textbf{Regions}: dorsal striatum, thalamus, frontal gyrus 
\\[3.5mm]

\cite{li2021braingnn} & Biopoint (118) & DK84 & Pooling (ROI) & 
\textbf{Regions}: frontal gyrus, temporal lobe, cingulate gyrus, occipital pole, angular gyrus 
\\[3.5mm]

\cite{yang2022identification} & ABIDE (303) & SF200 & Pooling (ROI) & 
\textbf{Regions}: right parietal cortex, left visual cortex, right lateral PFC, left PFC, left cingulate cortex
\\[3.5mm]

\cite{chu2022resting} & ABIDE (351) & AAL116 & Attention (ROI) & 
\textbf{Regions}: hippocampus, PARAH, putamen, thalamus \\
[3.5mm]

\cite{zhao2022multi} & ABIDE (92) [d] & AAL116 & Pooling (ROI) & 
\textbf{Regions} (Lo): bilateral cerebellum and right hippocampus ; 
\textbf{Regions} (Ho): left insula, left putamen, medial aspect of right SFG \\[3.5mm]

\cite{noman2022graph} & ABIDE (144) [d] & Power264 & Clustering (module) & 
\textbf{Modules}: ASD stronger FC in visual, DMN and SN ; TDC higher FC in sensory and auditory networks. \\[3.5mm]

\bottomrule
\end{tabularx}
\label{tab:asd1}
\end{table}

\begin{table}[ht]
\caption{Summary of findings from ASD studies that identified potential biomarkers. `Size' refers to the size of the dataset. When modalities beyond sFC are used, they are marked with [d] (dFC) or [M] (multimodal). `Type' refers to the granularity of attributions.
}
\begin{tabularx}{\textwidth}{L{0.5}L{0.70}L{0.5}L{0.80}L{2.5}}
\toprule
\textbf{Reference} & \textbf{Dataset (Size)} & \textbf{Atlas} & \textbf{Attributor (Type)} & \textbf{Salient features} \\
\toprule
\multicolumn{5}{c}{\textbf{ASD (continued)}} \\
\midrule

%%%%%%

\cite{zheng2024ci} & ABIDE (1064) & AAL116 & Info bottleneck (connection, module) & 
\textbf{Connections}: right STG - (left PARAH, right PrCG, right PoCG), \textbf{Modules}: SMN, SMN-LN \\[3.5mm]

\cite{zheng2024bpi} & ABIDE (1064) & AAL116 & Prototype (connection) & 
\textbf{Connections}: right SFG (orbital part) - right orbitofrontal cortex, right PoCG - (left PoCG, right PrCG) \\[3.5mm]

\cite{ma2024identification} & ABIDE (714) & AAL116 & Ablation (module) & 
\textbf{Modules}: DMN \\[3.5mm]

\cite{zheng2024brainib} & ABIDE (1099) & AAL116 & Info bottleneck (connection) & 
\textbf{Connections}: right superior parietal gyrus - (vermis, left MOG, right orbitofrontal cortex, cerebellum) \\[3.5mm]

\cite{wang2024multiview} & ABIDE (355) & AAL116 & GradCAM (ROI) & 
\textbf{Regions}: MTG, MFG, Precuneus, STG, Lingual gyrus \\[3.5mm]

\cite{wang2024adaptive} & ABIDE (860) & CC200 & Weights (ROI) & 
\textbf{Regions}: STG, Precuneus, supramarginal gyrus, SPL, MFG, parahippocampal gyrus, angular gyrus, putamen, left pallidum \\[3.5mm]
%%%%%%

\cite{kong2024multi} & ABIDE (618) [d] & AAL90 & Attention (ROI, connection) & \textbf{Regions}: left putamen, bilateral insula;
\textbf{Connections}: left medial SFG - right medial STF, left orbitofrontal cortex (superior medial) - right orbitofrontal cortex (superior medial) ; left IFG (triangular) - left IFG (operculum), right IFG (triangular) - right IFG (operculum) \\[3.5mm]

\cite{fang2024bilinear} & ABIDE (312) [M] & AAL116 & Attention (ROI) & \textbf{Regions}: right posterior cingulate gyrus, right ACG, right paracentral lobule, vermis \\[3.5mm]

\cite{wang2024ifc} & ABIDE (871) [d] & Unclear & Elastic Net (Connection) & \textbf{Connections}: right rectus gyrus - right cerebellum, left paracentral lobule - right middle occipital cortex, left precuneus - left cerebellum \\[3.5mm]

\bottomrule
\end{tabularx}
\label{tab:asd2}
\end{table}

\begin{table}[ht]
\caption{Summary of findings from ASD studies that identified potential biomarkers. `Size' refers to the size of the dataset. When modalities beyond sFC are used, they are marked with [d] (dFC) or [M] (multimodal). `Type' refers to the granularity of attributions.
}
\begin{tabularx}{\textwidth}{L{0.5}L{0.70}L{0.5}L{0.80}L{2.5}}
\toprule
\textbf{Reference} & \textbf{Dataset (Size)} & \textbf{Atlas} & \textbf{Attributor (Type)} & \textbf{Salient features} \\
\toprule
\multicolumn{5}{c}{\textbf{ASD (continued)}} \\
\midrule

\cite{gu2025fc} & ABIDE (871) & HO110 & Pooling (ROI) & \textbf{Regions}: temporal gyrus, PARAH, frontal cortex, accumbens, central cortex, thalamus, brainstem, and caudate nucleus \\[3.5mm]

\cite{wang2024leveraging} & ABIDE (184) [d] & AAL116 & Lasso (connection) & \textbf{Connections}: left SMG, right ACG ; left angular gyrus - left middle orbitofrontal cortex \\[3.5mm]

\cite{wei2023autistic} & Private (138) [M] & AAL90 & Eigenvalues (ROI) & \textbf{Regions}: left middle orbitofrontal cortex, bilateral supplementary motor area, left PrCG, bilateral rolandic operculum
\\[3.5mm]

\cite{bian2024adversarially} & ABIDE (663) & AAL90 & Graph filtration (connection) & \textbf{Connections}: left amygdala - left ITG, left hippocampus - right PoCG, left thalamus - right MTG, left dorsal SFG - left insulation \\[3.5mm]

\cite{gu2024novel} & ABIDE (871) & HO110 & Pooling (ROI) & \textbf{Regions}: Temporal gyrus, PARAH, frontal cortex, accumbens, central cortex, thalamus, brainstem, caudate nucleus \\[3.5mm]

\cite{wang2023consistency} & ABIDE (307) & AAL116 & GradCAM (ROI) & \textbf{Regions}: Precuneus
\\[3.5mm]

\cite{wang2023plsnet} & ABIDE (1009) & AAL116 & Pooling (ROI) & \textbf{Regions}: Temporal pole (MTG, STG), cerebellum\\[3.5mm]

\cite{zheng2023dynbraingnn} & ABIDE (582) [d] & Unclear & Attention (modular connection) & \textbf{Modular connection}: LN - SMN\\[3.5mm]

\cite{xu2024contrastive} & ABIDE (989) & SF100 & Attention (connection) & \textbf{Connections}: connections between PFC, parietal lobe, cingulate cortex \\[3.5mm]

\cite{zhang2023gcl} & ABIDE (987) & AAL90 & Trainable mask (ROI) & \textbf{Regions}: right MTG, left fusiform gyrus, left MOG, right linugal gryus, left MTG, right precuneus
% , right MOG, right calcarine fissure 
\\[3.5mm]

\cite{zhang2023gcl} & ABIDE II (532) & AAL90 & Trainable mask (ROI) & \textbf{Regions}: right MTG, left cerebellum, left MOG, right SFG, right STG, right precuneus 
\\[3.5mm]

\cite{wang2023modularity} & ABIDE NYU (184) & AAL116 & Lasso & \textbf{Connections}: thalamus, MTG - cerebellum \\[3.5mm]

\cite{menon2023asdexplainer} & Private (114) & Unclear & SubgraphX & \textbf{Regions}: right frontal orbital cortex \\[3.5mm]
 
\cite{hu2021gat} & ABIDE (1035) & HO110 & Multiple (connection) & \textbf{Connections} right hippocampus - right frontal medial cortex, left frontal pole - right IFG, left PrCG
\\[3.5mm]

\bottomrule
\end{tabularx}
\label{tab:asd3}
\end{table}

ASD encompasses a wide range of impairments of varying severity, but key characteristics include persistent social impairments and repetitive, restricted behaviour including experiencing distress when routines are disrupted. These signs should occur to an extent that affects daily life.  
As a neurodevelopmental disorder, it often begins in early childhood, often affects intellectual development and it could persist into adulthood. It has been observed to be more prevalent in boys than girls. 
Existing diagnostic biomarkers are primarily genetic, stemming from de novo mutations \citep{abi2023candidate}. While early studies posited that autistic children have larger head sizes, this was found to be limited to a subgroup \citep{libero2016persistence}. Findings from functional neuroimaging have yet to converge to a clear consensus \citep{lord2020autism}.

In this review, 42 studies on ASD were included, as shown in Tables \ref{tab:asd}-\ref{tab:asd3}. A majority of these studies depend on the ABIDE dataset, which is a multi-site consortium mostly represented by Western data sources (more from the USA than Europe). Despite its prevalent use, major issues with the dataset include variations in exclusion criteria (e.g. some sites do not include females, and have varying fluid intelligence thresholds) and inclusion criteria  - most sites relied on the same instruments (Autism Diagnostic Observation Scale / Autism Diagnostic Interview-Revised) but some used it in combination with clinical judgment, which would be subjective by nature.
Overall, model performance is rather high with a mean accuracy of 76.1\% (mean dataset size of 661). Notably, this remained consistent across dataset sizes. Out of the best-performing studies, studies using connection profiles as features and both BG and PG seem to have contributed to the improved performance. 

Given the large number of studies, we will only discuss the most frequently occurring salient found across multiple studies.
At the level of regions, precuneus, superior temporal gyrus, thalamus, prefrontal cortex, and cingulate gyrus/cortex were implicated in more than 3 studies. 
At the level of connections, no exact matches were found across all studies. 
At the level of modules, DMN and SN were highlighted in multiple studies but no modular connections were found to be reproducible.

Despite the presence of potential biomarkers that are reproduced across multiple sites, these regions are rather broad, and future research could consider using more fine-grained atlases to arrive at more specific ROIs. Additionally, considering the contrast between the presence of multiple reproducible salient regions and the lack of any reproducible salient connections, future studies could consider identifying salient subgraphs to see if there are any reproducible brain circuits that could reconcile the lack of agreement at the level of connections \citep{cash2023altered}.

\subsection{Major Depressive Disorder}
\label{sec-mdd}

\begin{table}[ht]
\caption{Summary of findings from MDD studies that identified potential biomarkers. `Size' refers to the size of the dataset. When modalities beyond sFC are used, they are marked with [d] (dFC) or [M] (multimodal). `Type' refers to the granularity of attributions.
}
\begin{tabularx}{\textwidth}{L{0.5}L{0.70}L{0.5}L{0.80}L{2.5}}
\toprule
\textbf{Reference} & \textbf{Dataset (Size)} & \textbf{Atlas} & \textbf{Attributor (Type)} & \textbf{Salient features} \\
\toprule
\multicolumn{5}{c}{\textbf{MDD}} \\
\midrule

\cite{qin2022using} & REST-meta-MDD (1586) & DOS160 & CAM (ROI, module) &  
\textbf{Regions}: right dorsal ACC, right ventrolateral PFC, left IPL, left posterior insula. 
\textbf{Modules}: DMN, FPN, Cingulo-Opercular network. \\[3.5mm]

\cite{gallo2023functional} & Multiple (2498) & Multiple & Multiple (connection) & 
\textbf{Connections}: stronger inter-hemispheric thalamic connection across most datasets and attributors \\[3.5mm]

\cite{kong2022multi} & Private (218) & Multiple & Weights (connection) & 
\textbf{Connections}: right anterior corona radiata - left dorsolateral PFC \\[3.5mm]

\cite{jun2020identifying} & Private (75) & YEO114 & Gradients (connection) & 
\textbf{Connections} (reduced in EC): left dorsal PFC - left precentral ventral region, striate cortex, parietal medial region, IPL, PARAH \\[3.5mm]

\cite{kong2021spatio} & Private (277) [d] & Unclear & Unclear (ROI) & 
\textbf{Regions}: bilateral pallidum, right putamen, bilateral MFG, right PoCG, right Heschl gyrus, right caudate, right olfactory cortex, right IFG, triangular part. \\[3.5mm]

\cite{fang2023unsupervised} & REST-meta-MDD (681) [d] & AAL116 & Attention (connection, temporal) & 
\textbf{Connections}: cross-hemisphere connections within the insula and lingual gyrus ; lingual gyrus - calcarine sulcus ; \textbf{Temporal}: middle \& end of time series \\[3.5mm]

\cite{zheng2024ci} & REST-meta-MDD (1604) & AAL116 & Info bottleneck (connection) & 
\textbf{Connections}: left rectus - (left cerebellum, right cuneus, right paracingulate gyri, left lingual gyrus, left SFG (medial orbital)) \\[3.5mm]

\cite{zheng2024bpi} & REST-meta-MDD (1604) & AAL116 & Prototype (connection) & 
\textbf{Connections}: left fusiform gyrus - right fusiform gyrus, left SPG - right SPG \\[3.5mm]

\cite{lee2024spectral} & REST-meta-MDD (470) & HO112 & t-test (ROI) & 
\textbf{Region}: ITG, frontal medial cortex, subcallosal cortex, PHG, temporal fusiform cortex, occiptal gyrus, brainstem \\[3.5mm]

\bottomrule
\end{tabularx}
\label{tab:mdd}
\end{table}

MDD presents as a state of persistently low mood (at least two weeks), often accompanied by andohenia. Other symptoms include feelings of worthlessness, suicidal thoughts, and poorer concentration. Significant changes in sleep, appetite, and activity levels often occur too. Existing fMRI studies have revealed 
significant differences between male and female patients \citep{tian2024gender}
and that regions in DMN, SN, and CEN are highly involved \citep{bondi2023systematic}. 
Specifically, DMN-linked connections include PCC - precuneus ; SN regions include ACC ; CEN connections include DLPFC - PCC, IFG-DMPFC. 
However, symptoms of MDD overlap with other disorders, making it difficult to identify biomarkers and traits unique to it. 

20 studies on MDD were reviewed, as shown in Table \ref{tab:mdd} and \ref{tab:mdd2}. A majority of them (16/20) used subsets of the REST-meta-MDD dataset, which is a collection of over 2000 scans from 17 hospitals in China. Other datasets include psymri (multiple sites, predominantly in Europe) and private datasets collected in China and Korea. Like other multi-site studies, the profile of patients varies widely across aspects such as severity, use of medication, and episodicity. 
Overall, model performance is moderately high (mean accuracy of 73.6\%, mean dataset size of 1003). Notably, smaller datasets (around 500 samples) exhibited greater variability of results. However, mean performance across dataset sizes (500, 1600, 2400) was rather consistent (around 74\%).

\begin{table}[ht]
\caption{Summary of findings from MDD studies that identified potential biomarkers. `Size' refers to the size of the dataset. When modalities beyond sFC are used, they are marked with [d] (dFC) or [M] (multimodal). `Type' refers to the granularity of attributions.
}
\begin{tabularx}{\textwidth}{L{0.5}L{0.70}L{0.5}L{0.80}L{2.5}}
\toprule
\textbf{Reference} & \textbf{Dataset (Size)} & \textbf{Atlas} & \textbf{Attributor (Type)} & \textbf{Salient features} \\
\toprule
\multicolumn{5}{c}{\textbf{MDD (continued)}} \\
\midrule

%%%%%%

\cite{zheng2024brainib} & REST-meta-MDD (1604) & AAL116 & Info bottleneck (connection) & 
\textbf{Connections}: left MTG - (right rolandic operculum, left fusiform gyrus, left PARAH, right SFG (medial orbital)) \\[3.5mm]

\cite{kong2024multi} & Zhongda (520) [d] & AAL90 & Attention (connection, ROI) & 
\textbf{Connections}: bilateral caudate, thalamus, paracentral lobule, posterior cingulate gyrus, cuneus, SOG ; left IFG triangular part - opercular part, right PrCG - right PoCG ; \textbf{ROI}: bilateral lingual gyrus, PoCG, PrCG, SOG ; right cuneus, left calcarine fissure \\[3.5mm]

\cite{kong2024multi} & REST-meta-MDD (667) [d] & AAL90 & Attention (connection, ROI) & 
\textbf{Connections}: bilateral caudate, insula, PoCG, SMG, SPG, IOG, SOG, left IFG triangular part - opercular part, right IFG triangular part - opercular part ; \textbf{Regions}: left posterior cingulate gyrus, right fusiform gyrus, right ITG, left MOG, left lingual gyrus, left calcarine fissure \\[3.5mm]

\cite{liu2024fusing} & REST-meta-MDD (533) & AAL116 & Pooling (ROI) & 
\textbf{Regions}: Precentral gyrus, SFG, Cuneus, Lingual gyrus, and Fusiform gyrus \\[3.5mm]

%%%%%%

\cite{gu2025fc} & REST-meta-MDD (533) & HO110 & Pooling (ROI) & 
\textbf{Regions}: left pallidum, right anterior ITG, left frontal orbital, right PARAH, left thalamus \\[3.5mm]

\cite{zhao2024enhancing} & REST-meta-MDD (1611) [d] & AAL116 & Attention (ROI, modules) & 
\textbf{Regions}: bilateral rectus, MTG, right middle frontal gyrus (orbital part)  ; \textbf{Modules}: DAN, FPN ; LN, DMN \\[3.5mm]

\cite{gu2024novel} & REST-meta-MDD (533) & HO110 & Pooling (ROI) & 
\textbf{Regions}: pallidum, ITG, frontal operculum cortex, PARAH, thalamus, temporal fusiform cortex, amygdala, accumbens and orbital cortex \\[3.5mm]

\cite{zhang2023slg} & REST-meta-MDD (2361) & AAL116 & Subgraph (ROI) & 
\textbf{Regions}: Thalamus \\[3.5mm]

\cite{kong2023multi} & Private (187) [M] & AAL116 & Weights (ROI) & 
\textbf{Regions}: bilateral medial SFG, bilateral dorsolateral SFG, bilateral caudate nucleus, bilateral precuneus, right hippocampus, and left lenticular nucleus \\[3.5mm]
%putamen

\cite{dai2023classification} & REST-meta-MDD (615) & AAL116 & GradCAM (ROI) & 
\textbf{Regions}: left insula, right amygdala, left SPG \\[3.5mm]

\cite{zhao2022detecting} & REST-meta-MDD (2361) & AAL116 & Pooling (ROI, module) & 
\textbf{Regions}: left inferior parietal lobe, right MFG, left PrCG
; \textbf{Modules}: DMN, LN \\[3.5mm]

\cite{wang2023modularity} & REST-meta-MDD (533) [d] & AAL116 & Lasso (connection) & 
\textbf{Connections}: hippocampus-IFG, cerebellum-support motor, thalamus-temporal pole/MTG \\[3.5mm]

\bottomrule
\end{tabularx}
\label{tab:mdd2}
\end{table}

Potential MDD biomarkers highlighted in these studies were found to have limited robustness. No salient features were consistently represented in the majority of the sites. This could be a result of the heterogeneous nature of the disorder.
Some potential biomarkers were present in multiple sites, though they tend to be present in only a few and are mostly limited to ROIs. This includes the pallidum, lingual gyrus, precentral gyrus, SFG, and thalamus. 
At the level of connections, none were found to be present in multiple studies. However, \cite{kong2024multi} highlighted several connections that were consistent across datasets (same model architecture, same attributor): left caudate - right caudate, left postcentral gyrus - right postcentral gyrus, left SPG - right SPG, left SOG - right SOG, right precentral gyrus - right postcentral gyrus, left IFG (triangular part) - left IFG (opercular part). 
At the level of modules, the default mode network and limbic network were most consistently highlighted across studies. 

In summary, these consolidated findings match past meta-reviews to a limited extent, with the greatest correspondence at the level of modules (DMN, SN), followed by ROIs (IFG). However, no matches at the level of connections could be found.

\subsection{Schizophrenia}
\label{sec-sz}

\begin{table}[ht]
\caption{Summary of findings from SZ studies that identified potential biomarkers. `Size' refers to the size of the dataset. When modalities beyond sFC are used, they are marked with [d] (dFC) or [M] (multimodal). `Type' refers to the granularity of attributions.
}
\begin{tabularx}{\textwidth}{L{0.5}L{0.70}L{0.5}L{0.80}L{2.5}}
\toprule
\textbf{Reference} & \textbf{Dataset (Size)} & \textbf{Atlas} & \textbf{Attributor (Type)} & \textbf{Salient features} \\
\toprule
\multicolumn{5}{c}{\textbf{SZ}} \\
\midrule

\cite{lei2022graph} & Multiple (1412) & Multiple & CAM (ROI) & 
\textbf{Regions}: decreased nodal efficiency in bilateral putamen and pallidum in SZ, across both atlases \\[3.5mm]

\cite{chen2023discriminative} & Private (345) [M] & Multiple & Pooling (ROI) &
\textbf{Regions}: bilateral rectus gyrus, bilateral lingual gyrus, bilateral cuneus; right medial orbitofrontal cortex, medial SFG, calcarine cortex, ACG\\[3.5mm]

\cite{sebenius2021multimodal}  & COBRE (154) [M] & DK293 & Pooling (ROI) & 
\textbf{Regions}: ROI relevance scores had a stronger correlation with SC than FC (specific regions unclear from plot) \\

\cite{zheng2024ci}  & SRPBS (184) & AAL116 & Information bottleneck (connection) & 
\textbf{Connections}: left hippocampus - left PARAH, left hippocampus - right hippocampus \\

\cite{zheng2024bpi}  & SRPBS (184) & AAL116 & Prototype (connection) & 
\textbf{Connections}: right ACG - right orbital part of MFG, left rectus - right orbital part of MFG, left orbital part of MFG - right orbital part of MFG ; left STG - left insula, left SMG - left insula, left putamen - left insula\\

\cite{zhu2024temporal}  & COBRE (112) [d] & AAL90 & Pooling (ROI, module) & 
\textbf{Regions}: left SOG, left fusiform gyrus, left medial SFG, left calcarine fissure, left IOG, left PrCG ; \textbf{Modules}: VN, DMN, DAN \\

\cite{zhu2024temporal} & UCLA (80) [d] & AAL90 & Pooling (ROI, module) & 
\textbf{Regions}: left lingual gyrus, left medial SFG, left IOG, left PrCG, left calcarine fissure ; \textbf{Modules}: DMN, VN, DAN \\

\cite{wang2024optimizing} & SRPBS (647) & BV140 & GNNExplainer (ROI) & 
\textbf{Regions}: ventricle, temporal gyrus \\

\cite{sunil2024graph}  & UCLA (177) & Custom AAL164 & GNNExplainer (ROI) & 
\textbf{Regions}: Supramarginal Gyrus (anterior division), ITG (posterior, temporooccipital, anterior division), STG (posterior division, left), right SPL, MTG (temporooccipital part, right) \\

\cite{li2023classification}  & In-house (143) [M] & AAL116 & GradCAM, Saliency (ROI) & 
\textbf{Regions}: bilateral insula, MFG, lower FG, MOG, angular gyrus, superior marginal gyrus \\

\cite{fan2023dgst}  & Multiple (1034) [d] & AAL116 & Attention (module) & 
\textbf{Module}: subcortical-cerebellar circuit \\

\bottomrule
\end{tabularx}
\label{tab:sz}
\end{table}

The definition of schizophrenia has evolved significantly over the past decades \citep{jablensky2010diagnostic}. There is a lack of consensus about how to name it \citep{lasalvia2018words} and there is still no clear definition even today.
At present, it requires both `positive' symptoms (such as delusional thinking, and auditory hallucinations) and `negative' symptoms (such as catatonic behavior, and social withdrawal) to be observed over time (at least a month) ; these symptoms have to be at an extent that affects one's daily life \citep{tandon2013definition}. Cognitive impairment is also prevalent in schizophrenia patients \citep{kahn2013schizophrenia}. Diagnosis is complicated by how symptoms might not show up all at once - it requires a prolonged period of monitoring that involves gradually excluding other related disorders. This motivates the search for alternative ways of diagnosing it that would require less time and therefore allow earlier intervention.

One view of schizophrenia is that of disruptions involving information processing. Functional imaging studies have revealed abnormalities in (i) salience processing (associated with positive symptoms) implicating the ventral striatum and DLPFC ; (ii) reward processing (associated with negative symptoms) implicating the amygdala, medial PFC as well as lowered ventral striatal responses to reward ; (iii) cognitive function (regulation and control of cognitive processes) implicating the DLPFC, rostral ACC and IPL \citep{Kahn2015} as well as lower connectivity between cortical (PFC) and subcortical regions (basal ganglia, thalamus, cerebellum) \citep{sheffield2016cognition}.

% dataset
In this review, a wide range of datasets were used by 11 studies on SZ, as shown in Table \ref{tab:sz}. This includes COBRE (USA), SRPBS (Japan), UCLA (USA), and various sites in China (AMU, FMMU, PKU, UEST). Most sites reported that diagnoses were arrived at via the Structured Clinical Interview used for DSM Disorders (SCID), but inclusion criteria for healthy controls varied.
Overall, model performance was high (mean accuracy of 85.4\%). We note that 4 sites had a significant class imbalance. However, after excluding them, the performance remained high (84.8\%). The average dataset size is 403 but we note that the performance is consistently high across small (around 100) and large datasets (above 1000, combination of multiple sites). This gives greater confidence to the salient features highlighted in these studies.

% biom
While there are no salient features that are consistently highlighted across all studies, there are some notable regions that are reported across multiple studies: medial SFG (Chen, Zhu UCLA, and COBRE), left lingual gyrus, left calcarine fissure (Chen, Zhu UCLA), anterior cingulate gyrus (Chen, Zheng), MFG (Zheng, Li). Such common findings are notably from different datasets (and countries), suggesting the possible presence of generalisable biomarkers across geographies.

% suggestions
Additionally, we note that \cite{zhu2024temporal} found a set of consistent ROIs across 2 datasets (COBRE, UCLA), including left medial SFG, left IOG, and left precentral gyrus. On the other hand, Zheng et al. arrived at quite different biomarkers on the same dataset when different novel attributors were used. This could suggest that attributors have a significant influence and that novel attributors should be carefully tested for robustness before they can be used.

Linking back to the pre-existing knowledge about the neural correlates of Schizophrenia, we found matches for positive symptoms and cognition-related regions but less so for negative symptoms.
A group of vision-related substructures such as lingual gyrus, calcarine fissure, and inferior occipital gyrus were found to be implicated, likely related to positive symptoms (e.g. visual hallucinations).
For cognition-related symptoms, we found that multiple Chinese sites identified ACC to be important. However, this was not replicated in Western studies (e.g. UCLA, COBRE).
Finally, the relatively consistent findings of the medial superior frontal gyrus being implicated might give a more specific localisation of DLPFC-related aberrations in FC. However, it is unclear from the studies whether this is in any way more strongly identified in specific subgroups of schizophrenia symptoms.

\subsection{Transdiagnostic biomarkers}
\label{transdx}

Psychiatric disorders often have overlapping symptoms and numerous studies have argued for them to be studied in tandem \citep{van2019cross,de2019shared}. 
Having summarised the reproducible salient features for each disorder, another round of matching was performed to identify any overlaps of these potential biomarkers across disorders. This results in the following matches:
\begin{itemize}
    \item Medial frontal gyrus (ADHD and SZ)
    \item Cingulate gyrus (ASD and SZ) 
    \item Lingual gyrus (MDD and SZ)
    \item Thalamus (ASD and MDD)
    \item DMN (ASD and MDD)
    \item SN (ASD and MDD)
\end{itemize}

It is clear that SZ has significant overlaps with multiple disorders (ADHD, ASD, MDD), in terms of common brain regions being affected. 
Additionally, the overlap between ASD and MDD seems stronger than other combinations, with multiple common brain modules and regions being affected in both disorders. 
% However, these regions represent rather large areas of the brain an

One caveat to this analysis is the possibility that other overlapping brain regions are not considered (e.g. due to the limited number of studies in this review for disorders like ADHD). Nevertheless, this shortlist of regions was found to exhibit a greater degree of robustness across studies (i.e. across pre-processing pipelines, choice of predictor and attributor, etc.) and could be explored in future studies.

\subsection{Summary}

Linking back to the third question mentioned in the introduction (whether any salient features are reproducible, whether any transdiagnostic biomarker is present), we have identified a list of potential biomarkers that are present across multiple studies of the same disorder and even identified several potential transdiagnostic biomarkers. However, these are limited to brain regions. Functional connections reported in individual studies exhibited very poor robustness with minimal reproducibility across studies. 

Overall, studies on schizophrenia produced the highest classification performances, suggesting that it is possible for GNNs to distinguish SZ patients from healthy controls via fMRI and that the potential biomarkers identified could be more reliable (than those found in studies of other disorders).
Research on the other disorders (ASD, ADHD, MDD) could benefit from more thorough benchmarking of state-of-the-art GNN models, or from pivoting to a subtyping approach (i.e. splitting the patient population into subgroups could make it more feasible for disorder classification models to perform better and produce more reliable salient features).

\section{Discussion}
\label{sec:disc}

\begin{figure}[h]
	\centering
		\includegraphics[width=0.88\linewidth]{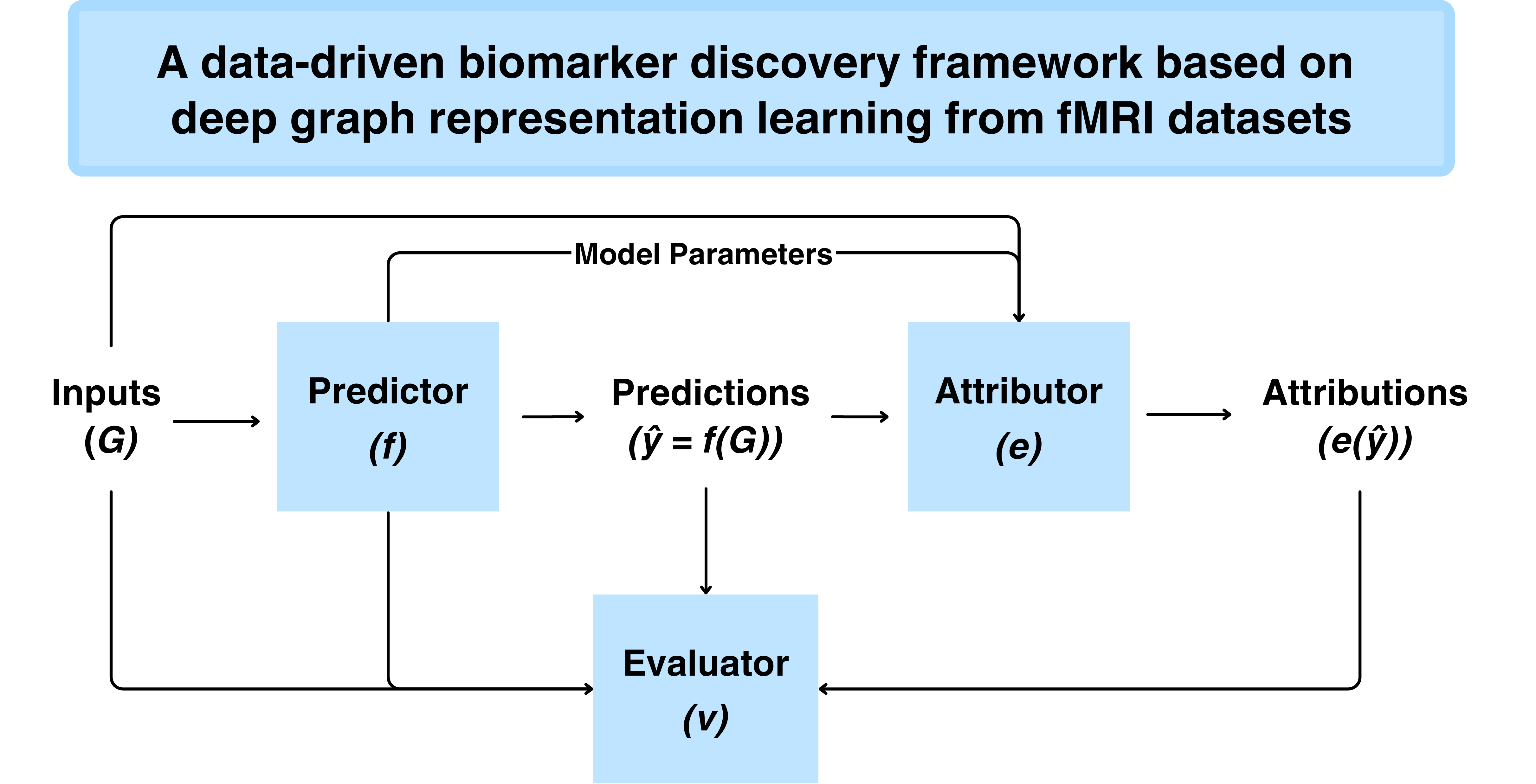}
	\caption{An illustration showing where the three key stages occur in a typical pipeline that uses a post-hoc attributor. For intrinsically interpretable models, the attributor would become part of the predictor (e.g. pooling layers can be used for producing explanations, but they form part of the GNN architecture).}
	\label{fig:pipeline}
\end{figure}

In order to discover robust biomarkers of psychiatric disorders via machine learning techniques, it is necessary for each component of the biomarker discovery pipeline to work well: (i) a high-performing \textbf{predictor} that generalises well out of sample, (ii) an \textbf{attributor} that produces stable attribution scores robust to changes unrelated to the disorder, (iii) an \textbf{evaluator}, consisting of a set of metrics to evaluate the predictor-attributor combination, that provides objective evidence that the scores corroborate with existing understanding of the disorder. Figure \ref{fig:pipeline} provides a visualisation of how these 3 components work together.

In this review, we have provided an analysis of the latest research in all 3 components and identified potential biomarkers for several psychiatric disorders, including transdiagnostic ones. Summarising the answers to the questions presented in the introduction, we found that the best predictors tend to use thresholded FC matrices as the adjacency matrix, connection profile as node features and involve both BG and PG. Some graph-based attributors such as GNNExplainer provide greater flexibility of attribution targets than generic ones. The use of evaluators to ascertain the robustness of proposed biomarkers has much room for improvement as they are currently limited to a minority (about 10\%) of existing studies. In our own analysis of the reproducibility of these biomarkers, region-level features were found to exhibit greater robustness than edge-level features and this is consistently observed across multiple disorders. 
In the following discussion, we will provide a deeper evaluation of each component.

\begin{figure}[h]
	\centering
		\includegraphics[width=0.88\linewidth]{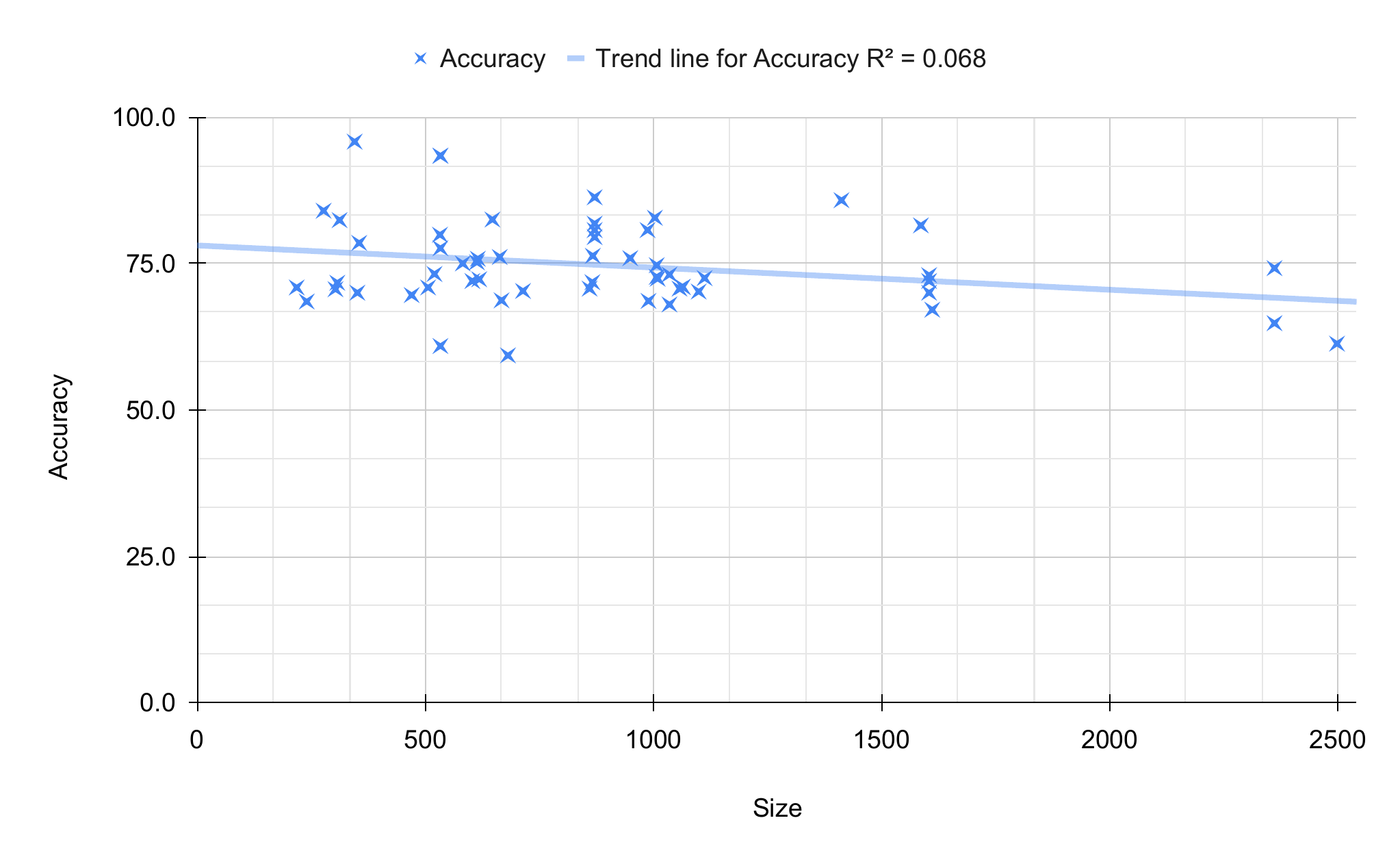}
	\caption{Plot of model accuracy against dataset size. It is evident that a majority of the studies have dataset sizes below 1000 and classification accuracies generally decrease as the dataset size increases.}
	\label{fig:size}
\end{figure}

We found the reported classification performances of \textbf{predictors} to be generally high. 
However, there are several caveats to this finding. First, most studies do not account for confounds such as age and sex, which could have inflated the model performance (e.g. model captures sex-specific features that will fail to generalise to other datasets). We note that one way to address this is to ensure that these factors are balanced across classes (healthy/disorder), but doing so would often reduce dataset sizes further. On the other hand, the flexibility afforded by GNNs via the use of PGs to incorporate such demographic information could help to alleviate this (e.g. existing BG-only GNNs could add on a PG to account for these factors as covariates).
Second, most studies do not demonstrate generalisation to a separate dataset not used during training (i.e. their results merely show their best performance on a test data split from the same dataset, but often do not go further to show generalisation capabilities to a completely different dataset).
Third, a deeper analysis reveals that such good performance might be boosted by studies that rely on small datasets. 
A previous study by \citep{teng2023brain} has demonstrated that classification accuracy drops when dataset size increases. This phenomenon is also observed in our analysis, as shown in Figure \ref{fig:size}. This suggests that simply changing the architecture (e.g. to GNN) might not lead to better model performance. Rather, other factors such as dataset size (specifically, the extent of heterogeneity in the sample) have a greater effect on model performance and solutions specific to these issues (rather than just using the latest variant of neural network architectures) need to be developed.

In view of this, we reiterate the need for proper benchmarking. 
Although several benchmarking studies have been done using GNNs on fMRI datasets, they were focused on baseline GNNs and not state-of-the-art models that have been shown to do better than baselines. 
Several considerations need to be taken care of in future benchmarking studies: (i) use datasets that are sufficiently large and share precise information about how the dataset is split, so that future studies can follow them, (ii) use of state-of-the-art GNN models, (iii) control for size of model (i.e. number of trainable parameters should be kept similar across models), (iv) report performance metrics beyond just accuracy (e.g. AUC) so as to account for class imbalances, (v) demonstrate generalisability beyond the dataset used for training (i.e. to a new dataset, not just the test split).  

Nevertheless, we caution against dismissing a model just because it generalised to unseen data poorly. 
Intuitively, a model that has limited performance degradation when tested on unseen datasets would suggest that the features identified by the model to be salient are indeed also important in other datasets of the same disorder.
However, more care is needed when choosing these unseen datasets.
For instance, datasets aggregated from multiple sites (i.e. data consortiums) are increasingly popular. Such datasets are often gathered from different countries and poor generalisation performance does not immediately mean that fMRI is not useful for identifying patients with the disorder. 
Rather, it could indicate the presence of heterogeneity in the disorder, especially if the model works well on a test split of the same dataset but performance decays rapidly when using a different dataset (or different imaging site). Such models could still accurately capture site-specific biomarkers that are possibly characteristic of a subtype of the disorder.
Thus, more careful evaluation is needed, e.g. using a different site with similar demographics and inclusion/exclusion criteria as the test set. 
On the other hand, if the goal is to discover biomarkers that are generalisable across the population, then sizable datasets should be expected to be used to justify such claims and demonstration of limited performance degradation in out-of-sample distributions should become a requirement.

On a final note for predictor-related issues, it could be more meaningful to move beyond the use of class labels, especially in view of the litany of studies on disorder classification and their limitations.
Binary classification serves as a simplified way to study disorders since patients can be contrasted against `healthy' / `normal' controls.
However, such class labels are a preliminary construct guided by our existing imperfect understanding of these disorders. 
Furthermore, datasets aggregated from multiple sites are often used without paying heed to variations in inclusion criteria, blurring the lines between what is defined as a control subject and a patient.
Thus, it might not be the best approach to constrain our learning algorithms with these labels, especially when the goal is to go beyond our current understanding. 
For instance, psychiatric disorders have a large number of possible combinations of symptoms that lead to the same diagnosis even though they have different underlying biology \citep{chen2023leveraging}. In these scenarios, class labels could be unreliable, especially for poorly understood disorders.
On the other hand, using a completely data-driven approach is very challenging. For instance, clustering-based approaches could be applied directly to the data but cluster interpretation is often fraught with subjectivity.
Instead, one could predict test scores (such as ADOS, HAM-D, PANSS) or even imaging-guided labels such as PET grading \citep{li2022predicting}. 
This will require the use of regression instead of classification, but the majority of attributors are also able to produce attribution scores in such models.

In terms of the use of \textbf{attributors}, it is clear from this review that the choice of attributor restricts the granularity of salient features that can be highlighted. GNN-specific explainers have the distinct advantage of producing subgraph-level attributions. In view of the lack of agreement across edge-level attributions, subgraph/motif-level attributions could be a promising direction as alluded to in \cite{cash2023altered}, where they showed how a circuit-level analysis can unify seemingly contradictory findings at lower granularities. 

Another key insight is that even within the same category, the choice of attributor could introduce significant variability to attribution scores, even when the dataset and predictor remain the same.
This warrants a separate benchmarking study to determine the best combination of predictors and attributors for biomarker discovery. 
In spite of these instabilities, we have identified several consistent region-level biomarkers for each disorder, as well as candidate transdiagnostic biomarkers. However, there are several caveats to these findings.

Firstly, most studies with large datasets almost always rely on data consortiums that collect data from multiple sites (so as to address issues of small datasets, such as inflated accuracies and high standard deviations).
While some data consortiums try to standardise scanning protocols to minimise scanner-induced variabilities, the majority of such consortiums pool datasets without standardisation. 
Existing studies have demonstrated that salient features identified from such aggregated datasets tend to be biased towards the largest site \citep{chan2022semi} and the use of data harmonisation algorithms like ComBat could lead to changes to salient features \citep{chan2023elucidating}. 
Thus, better data harmonisation techniques would need to be developed and these changes would require further study before they can be used for biomarker discovery.

Secondly, a major limitation of existing studies is that salient features are often aggregated at a class-wide level. However, for them to be clinically useful, attributors that can produce individualised attributions should be used (e.g. IG ; global attributors such as XGNN would not be useful in such a situation).  
However, much work is still needed to study the consistency of these attributors before they can be used for elucidating individual insights.

Thirdly, one major challenge faced when assessing the robustness of the potential biomarkers brought up by these studies is the lack of consistency in the way the most salient features are reported.
This problem is caused by three factors: (i) different atlases used, (ii) unavailability of the raw attribution scores and a lack of clarity on whether biomarkers reported are averaged across test data only, or the entire dataset, (iii) different extent of thoroughness in reporting the salient features (e.g. listing top 10/20/30 features, or just mentioning a few salient features in passing).
Furthermore, different granularities of the biomarkers are available due to the design of the GNN architecture and the choice of attributors makes it difficult to harmonise certain features that are less often reported.
In this study, we found that a majority of the studies (64.1\%) report ROI-level features while another significant portion of studies reports at the level of connections (24.4\%). However,  functional brain modules (6.4\%) and modular connections (5.1\%) are less often reported.
Overall, harmonising these findings manually is infeasible at a large scale without improvements in reporting standards.

To address this, we suggest the following guidelines as a preliminary step toward establishing a standard that future publications can refer to:
\begin{itemize}
    \item Attribution scores for each feature should be recorded in a spreadsheet that is shared along with the publication (e.g. as supplementary materials). Minimally, they should contain the ranking of the features, or be sorted according to importance. Ideally, if scores are available for each individual subject, they should be shared too (e.g. as NumPy files). 
    \item Key metadata about the most salient features (e.g. ROI names as defined by the atlas, along with their MNI coordinates) should be provided in the supplementary materials in the form of a spreadsheet, so as to facilitate future research and meta-analysis.
    \item Whenever attributors provide information about polarity (i.e. positive and negative scores, reflecting hyper/hypo-connectivity \citep{gupta2022decoding}), these raw values should be reported even if only the absolute value / magnitude is used in the analysis.
\end{itemize}

Having such guidelines would make it possible to develop computational tools to automate meta-analysis, helping research in this area to progress more quickly.

In the case of \textbf{evaluators}, the use of evaluation metrics to measure the robustness of these attribution scores has been an emerging trend in recent years, but little has been done to determine what metrics are appropriate and required for biomarker discovery. In this review, we have attempted to identify the most relevant subset of the Co-12 framework, but much more remains to be done to assess the robustness of the attribution scores produced by attributors. For instance, several unknowns remain, such as an appropriate number of features to keep/remove in checks for Completeness. 
Once resolved, such scores would provide another set of information, on top of model prediction performances, to determine the robustness of the proposed architecture.  

More crucially, the generic evaluation metrics merely serve as a starting point for this research direction. 
Many novel evaluation metrics could also be proposed in future studies to address other desiderata not covered by these metrics. One example used in radiology applications is the prediction-saliency correlation metric \citep{zhang2023revisiting} which computed the correlation between changes in model predictions and the corresponding saliency maps.
Furthermore, metrics that capture brain-specific information could be proposed. Some examples include modularity ratio and hub assortativity coefficient as proposed in \citep{girishrobustness}.

Overall, there are several limitations to the findings we have arrived at. 
One aspect not covered in our analysis is the effect of variability in pre-processing pipelines (e.g. whether to perform global signal regression, Fisher transform, etc.) on model performance and biomarker robustness. 
These issues are complex and should be analysed in a separate study. 
One way to reduce the variability of pre-processing pipelines is to release more pre-processed connectome datasets. 
We note from our analysis that the availability of pre-processed datasets released by data consortiums drives research: about 1/2 of the studies are on ASD and almost all of them used the ABIDE I dataset, especially the preprocessed connectomes with 871 subjects. 
While the ABIDE II dataset is also available, few studies utilised it due to the lack of pre-processed data.
Having pre-processed data made available (on top of the raw brain images) helps to remove one source of variability and also allows results to be compared across studies. 
Thus, we encourage future data releases to also share pre-processed fMRI datasets or even connectomes. 

Additionally, the scope of this review paper is limited to studies that included resting-state fMRI and it is important to keep in mind that it only provides information about brain function at rest. Whenever possible, other modalities (structural, molecular, task-fMRI, etc.) should be studied in tandem to produce a complete picture.
In this regard, GNN provides a very flexible architecture to model multi-modal datasets: (i) modalities that are traditionally image-based (e.g. T1w, PET) can still be converted into graphs \citep{zhang2023multi} such as morphometric similarity networks \citep{sebenius2021multimodal,park2023graph},
(ii) GNN can be used to integrate connectome datasets with multi-omics data \citep{ghosal2023interpretable} in a parameter-efficient manner, 
(iii) population graphs make it possible to integrate non-imaging modalities \citep{parisot2018disease}. 
Multi-modal biomarkers will be another emerging area that would give a more holistic view than what is currently possible with studies only using fMRI \citep{chen2023leveraging}.

Finally, we note that our review does not include details about temporal GNNs, counterfactual-based explanations (such as \citep{shen2024gcan}) and causal discovery \citep{rawls2023causal}. These areas of research are exciting novel directions in biomarker discovery from rs-fMRI data as they go beyond current limitations of static FC, factual explanations and correlation-driven studies, respectively. 
However, each of these research areas is complex, is relatively less explored within the studies considered in our review, and would require careful exploration to determine their utility for biomarker discovery. For instance, temporal GNNs are essential for dFC-based biomarker discovery, but they face issues such as aliasing effects, determining the appropriate window and step size, etc. \citep{esfahlani2022edge}. These issues could have significant implications on the robustness of dFC biomarkers and have not been well-studied yet (for instance, evaluation metrics sensitive to temporal attributions are relatively under-explored). 
We refer interested readers to Kakkad \emph{et al.} \citep{kakkad2023survey} for a review on temporal GNNs and counterfactual explanations, and \citep{bielczyk2019disentangling} for a review on causal discovery from fMRI.

\section{Conclusion}

In summary, while there has been an abundance of novel GNN architectures designed for disorder prediction from fMRI data, there remains much room for further research to improve the robustness of the salient features highlighted by these predictors and attributors. 
Benchmarking studies that involve state-of-the-art GNN predictors customised for fMRI datasets are needed. Studies on optimal choices of predictors and attributors are also required as their robustness on fMRI datasets is still poorly understood.
Existing evaluation metrics were designed for generic (graph) datasets and more metrics appropriate for FC datasets are needed to determine whether attributors are sensitive to known properties of FC. 
The lack of standardised reporting of salient features has made it challenging to consolidate insights from existing studies. This is a complex issue that needs to be revisited once a better understanding of predictors and attributors is developed, and when better metrics have been created.
Finally, the paucity of reproducible salient features (especially at the level of connections) motivates the search for alternative approaches.
Possible directions for future research include looking beyond the use of class labels and pivoting away from the goal of solely chasing for generalisable biomarkers for the entire disorder class, especially for heterogeneous disorders. Moving towards regression tasks, transdiagnostic studies and more fine-grained biomarkers could result in more robust biomarkers of psychiatric (and more generally, neurological) disorders being discovered from fMRI datasets in the near future. 

\section*{Funding}
This work was supported in part by AcRF Tier-2 grant MOE T2EP20121-0003 of Ministry of Education, Singapore.

\newpage

\appendix
\section{Appendix}

\subsection{Rationale for exclusion of 4 Co-12 factors}
 
\begin{enumerate}
    \item \textbf{Controllability}: This concerns the extent a user has control over an explanation (e.g. ability to correct, or manipulate it within an interface). Being applicable only to interactive explanations, it might not have much relevance at present. However, it could be an important consideration if future clinical use of biomarkers involves clinicians interacting with a graphical user interface that allows them to explore various granularities of biomarkers personalised to the patient.  
    \item \textbf{Context}: This entails creating explanations that consider the users' requirements to ensure that they understand the explanations. 
    It is indeed a relevant consideration, but we do not think that biomarkers for psychiatric disorders are at a state where it is ready for clinical deployment. Once robust biomarkers are established, it would then be necessary to determine exactly which components of the biomarkers to present to address clinicians' needs, as well as the mode of presentation for such information in clinical settings. 
    \item \textbf{Composition}: This is another potentially relevant aspect that concerns the formatting and organisation of explanations. For instance, whether biomarkers should be presented at the level of nodes, edges, motifs/subgraphs, or within the entire connectome. However, at this stage, it is necessary to first ascertain which granularity of biomarkers is most robust and thus this aspect of Co-12 would be a greater concern when functional biomarker discovery for psychiatric disorders is at a more mature stage. 
    \item \textbf{Covariate complexity}: This relates to having human-understandable explanations, but FC studies typically use ROIs as features which are already understandable, unlike individual pixels in images.
\end{enumerate}

\subsection{Supplemental Tables}

In this subsection, several tables are presented for a convenient reference of the mapping between abbreviations used in this review paper to their full names.

\begin{table}[h]
\caption{Mapping of abbreviations used for atlases to their full names. More information about the atlases can be found in the corresponding papers that used the atlas in their study. Note that this list is limited to atlases used in studies considered in this review (e.g. it does not include Glasser, or variants of Schaefer's atlas.}
\label{tab:atlas}
\begin{tabular}{lll}
\hline
Atlas & Full name & ROIs \\
\hline

\hline
AAL116 & Automated anatomical labelling & 116 \\
AAL166 & Automated anatomical labelling v3 & 166 \\
AAL90 & Automated anatomical labelling & 90 \\
BASC325 & Bootstrap Analysis of Stable Clusters & 325 \\
BM82 & Broadmann & 82 \\
BN273 & Brainnetome & 273 \\
BNA246 & BrainNet Atlas & 246 \\
BV140 & BrainVISA Sulci & 140 \\
CC200/400 & Craddock & 200/400 \\
DK308 & Desikan–Killiany & 308 \\
DK86 & Desikan-Killiany, with subcortical ROIs & 86 \\
DOS160 & Dosenbach & 160 \\
DX148 & Destrieux & 148 \\
EZ115 & Eickoff‐Zilles & 115 \\
HO110/111/112 & Harvard-Oxford & 110/111/112 \\
JHU81 & JHU ICBM-DTI-81 & 81 \\
MODL128 & Dictionaries of functional modes & 128 \\
Power264 & Power atlas & 264 \\
SF200 & Schaefer & 200 \\
SHEN268 & Shen atlas & 268 \\
TT93 & Talairach and Tournoux & 93 \\
TT97 & Talariach & 97 \\
YEO114 & Yeo 17-network & 114\\
\hline
\end{tabular}
\end{table}

\begin{table}[h]
\caption{Mapping of common abbreviations used to their full names.}
\label{tab:abbrev}
\begin{tabular}{lll}
\hline
Abbreviation & Full name \\
\hline
1-WL & 1-dimensional Weisfeiler-Leman \\
ADHD & Attention Deficit Hyperactivity Disorder \\
ASD & Autism Spectrum Disorder \\
BG & Brain Graph \\
CAM & Class Activation Mapping \\
CNN & Convolutional Neural Network \\
dFC & dynamic Functional Connectivity \\
DNN & Deep Neural Network \\
FC & Functional Connectivity \\
fMRI & functional Magnetic Resonance Imaging \\
GAT & Graph Attention Network \\
GCN & Graph Convolutional Network \\
GIN & Graph Isomorphism Network \\
GNN & Graph Neural Network \\
HSIC & Hilbert-Schmidt Independence Criterion \\
ICA & Independent Component Analysis \\
IG & Integrated Gradients \\
k-NN & k-Nearest Neighbours \\
MDD & Major Depressive Disorder \\
MI & Mutual Information \\
ML & Machine Learning \\
NC & Normal Controls \\
NLP & Natural Language Processing \\
PCA & Principal Component Analysis \\
PG & Population Graph \\
RF & Random Forest \\
RFE & Recursive Feature Elimination \\
RL & Reinforcement Learning \\
ROI & Region of Interest \\
sFC & static Functional Connectivity \\
SNR & Signal-to-Noise Ratio \\
SVM & Support Vector Machine \\
SZ & Schizophrenia \\
TDC & Typically Developing Children\\
\hline
\end{tabular}
\end{table}

\begin{table}[h]
\caption{Mapping of abbreviations used for brain networks to their full names.}
\label{tab:mod}
\begin{tabular}{lll}
\hline
Abbreviation & Full name \\
\hline

\hline
DAN & Dorsal Attention Network\\
CEN & Central Executive Network\\
DMN & Default Mode Network\\
FPN & Frontoparietal Network\\
LN & Limbic Network\\
SMN & Somatomotor Network\\
SN & Salience Network\\
VAN & Ventral Attention Network\\
\hline
\end{tabular}
\end{table}

\begin{table}[h]
\caption{Mapping of abbreviations used for brain regions to their full names.}
\label{tab:roi}
\begin{tabular}{lll}
\hline
Abbreviation & Full name \\
\hline

\hline
ACC & Anterior Cingulate Cortex \\
ACG & Anterior Cingulate Gyrus \\
FOG & Frontal Orbital Gyrus \\
FP & Frontal Pole \\
IFG & Inferior Frontal Gyrus \\
IPL & Inferior Parietal Lobule \\
MFG & Middle Frontal Gyrus \\
MOG & Middle Occipital Gyrus \\
MTG & Middle Temporal Gyrus \\
PARAH & Parahippocampal Gyrus \\
PoCG & Postcentral Gyrus \\
PrCG & Precentral Gyrus \\
PFC & Prefrontal Cortex \\
SFG & Superior Frontal Gyrus \\
SMG & Superior Medial Gyrus \\
SOG & Superior Occiptal Gyrus \\
SPG & Superior Parietal Gyrus \\
STG & Superior Temporal Gyrus \\
TP & Temporal Pole\\
\hline
\end{tabular}
\end{table}

\newpage

\bibliographystyle{cas-model2-names}

\bibliography{cas-refs}

\newpage

\section*{Supplementary Materials}
\pagestyle{empty}

\begin{figure}[h]
	\centering
		\includegraphics[width=0.8\linewidth]{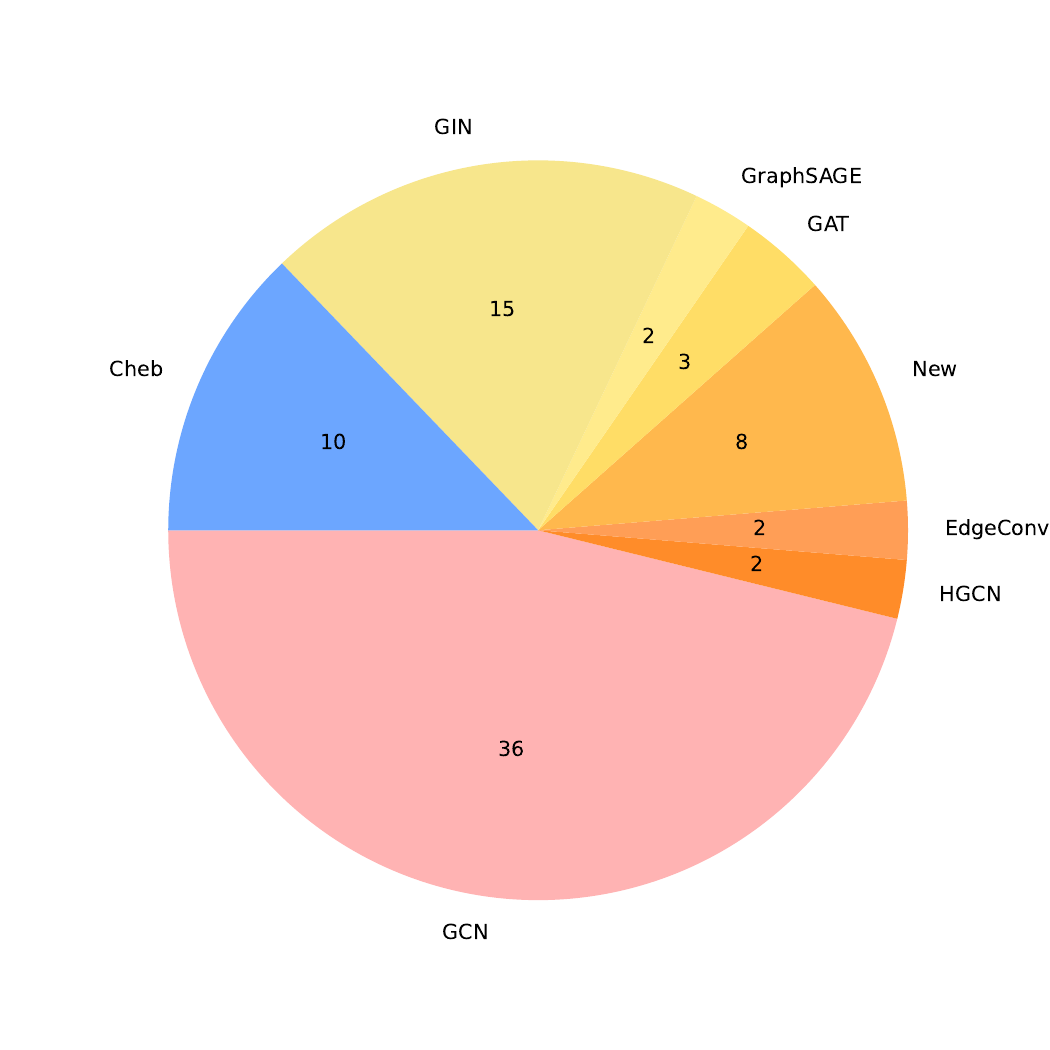}
	\caption{Choice of GNN used in the papers reviewed. Some studies included more than one type of GNN. `New' refers to architectures that significantly deviate from baseline GNNs (e.g. customised message passing techniques). Colour scheme represents various classes of GNNs - Light red: GCN (spatial/spectral), Shades of yellow: Spatial GNN, Blue: Spectral GNN, Shades of orange: Others).}
	\label{fig:pie_pred}
\end{figure}

\begin{figure}[h]
	\centering
		\includegraphics[width=0.8\linewidth]{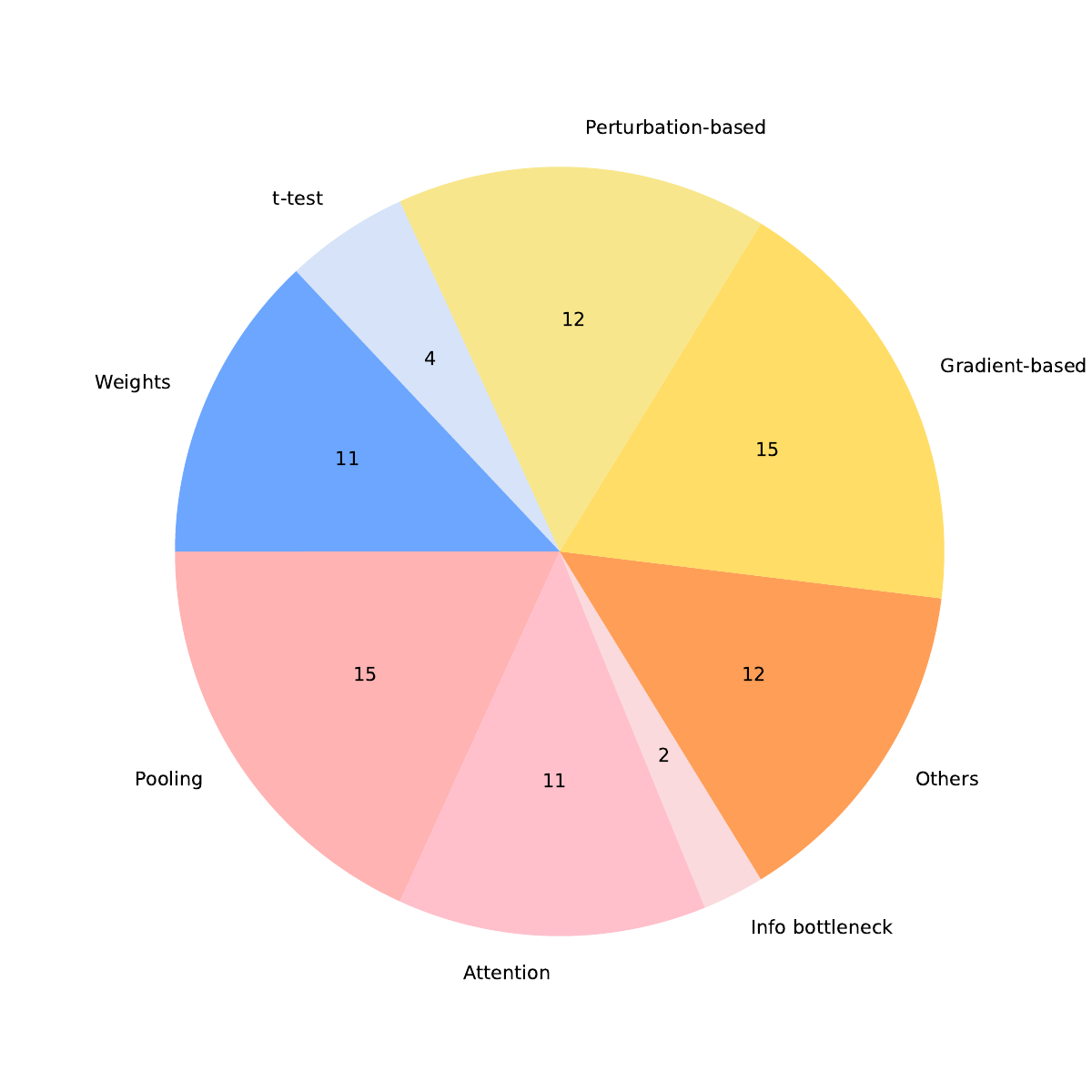}
	\caption{Distribution of attributors covered in our review. Some studies used more than one type of attributor. Colour scheme represents various categories of attributors, following our proposed taxonomy - Shades of red: Self-interpretable, Orange: Others, Shades of yellow: Post-hoc, Shades of blue: `Traditional' approaches.}
	\label{fig:pie_attri}
\end{figure}

\begin{table*}[h]
\caption{Summary of findings from ADHD studies, including those with biomarker discovery performed (placed above midline). `Size' refers to the size of the dataset. When modalities beyond sFC are used, they are marked with [d] (dFC) or [M] (multimodal).
}
\begin{center}
\begin{tabularx}{\textwidth}{@{} L{0.33}L{0.6}L{0.92}L{0.4}L{0.5}L{0.50}L{0.33}L{0.67} @{}}
\toprule
\textbf{Reference} & \textbf{Dataset (Size)} & \textbf{Dataset distribution} & \textbf{Atlas} & \textbf{GNN} & \textbf{Graph} & \textbf{Result} & \textbf{Baseline} \\
\toprule

\cite{yu2022graph} & CNI-TLC (240) & 120 ADHD, 120 TDC & CC200 & Kipf & BG & 68.5\% & GAT 58.6\% \\[.7mm]

\cite{zhao2022dynamic} & ADHD-200 (603) & 260 ADHD, 343 TDC & AAL116 & New & BG & 72.0\% & GAT 68.0\% \\[.7mm]

\cite{luo2024knowledge} & ADHD-200 (506) & 73 ADHD, 75 TDC & AAL116 & GCN & BG & 75.8\% & GCN 71.7\% \\[.7mm]

\cite{zhang2023gcl} & ADHD-200 (506) & 215 ADHD, 291 TDC & Multiple & GIN & BG & 70.9\% & SAGE 62.1\% \\ [.7mm]

\bottomrule
\end{tabularx}
\label{tab:supp-adhd}
\end{center}
\end{table*}

\begin{table*}[h]
\caption{Summary of findings from ASD studies, including those with biomarker discovery performed (top section), without biomarker discovery performed (middle section), and smaller datasets (bottom section). `Size' refers to the size of the dataset. When modalities beyond sFC are used, they are marked with [d] (dFC) or [M] (multimodal).
}
\begin{center}
\begin{tabularx}{\textwidth}{@{} L{0.33}L{0.65}L{0.9}L{0.4}L{0.5}L{0.50}L{0.33}L{0.67} @{}}
\toprule
\textbf{Reference} & \textbf{Dataset (Size)} & \textbf{Dataset distribution} & \textbf{Atlas} & \textbf{GNN} & \textbf{Graph} & \textbf{Result} & \textbf{Baseline} \\
\toprule

\cite{li2021identification} & ABIDE (866) & 402 ASD, 464 TDC & MODL128 & Defferrard & BG & 71.8\% & k-NN 65.8\% \\[.7mm]

\cite{zhang2022classification} & ABIDE (871) & 403 ASD, 468 TDC & HO112 & Defferrard & Mix & 81.8\% & pGCN 71.4\% \\[.7mm]

\cite{wang2022mage} & ABIDE (949) & 419 ASD, 530 TDC & Multiple & Kipf & PG & 75.9\% & MLP 75.2\% \\[.7mm]

\cite{zhu2021triple} & ABIDE (1112) & Unclear & AAL116 & GraphSAGE & BG & 72.5\%& pGCN 69.7\% \\[.7mm]

\cite{shao2021classification} & ABIDE (871) & 403 ASD, 468 TDC & HO111 & Kipf & PG & 79.5\% & MLP 78.1\% \\[.7mm]

\cite{wang2021graph} & ABIDE (1057) & 525 ASD, 532 TDC & CC200 & Edge & BG & 70.7\% & DNN 70.0\% \\[.7mm]

\cite{li2022te} & ABIDE (871) & 403 ASD, 468 TDC & Multiple & Defferrard & Mix & 86.3\% & pGCN 69.5\% \\[.7mm]

\cite{zhu2022contrastive} & ABIDE (613) [d] & 286 ASD, 327 TDC & HO110 & Mix & PG & 75.2\% & MVGCN 72.0\% \\[.7mm]

\cite{cui2023dynamic} & ABIDE (1035) [d] & 505 ASD, 530 TDC & CC200 & Edge & Mix & 73.1\% & SVM 64.0\% \\[.7mm]

\cite{chen2022invertible} & ABIDE (867) [d] & 416 ASD, 451 TDC & HO110 & Defferrard & BG & 76.3\% & GCN 73.2\% \\[.7mm]

\bottomrule
\end{tabularx}
\label{tab:supp-asd}
\end{center}
\end{table*}

\begin{table*}[h]
\caption{Summary of findings from ASD studies, including those with biomarker discovery performed (top section), without biomarker discovery performed (middle section), and smaller datasets (bottom section). `Size' refers to the size of the dataset. When modalities beyond sFC are used, they are marked with [d] (dFC) or [M] (multimodal).
}
\begin{center}
\begin{tabularx}{\textwidth}{@{} L{0.33}L{0.65}L{0.9}L{0.4}L{0.5}L{0.50}L{0.33}L{0.67} @{}}
\toprule
\textbf{Reference} & \textbf{Dataset (Size)} & \textbf{Dataset distribution} & \textbf{Atlas} & \textbf{GNN} & \textbf{Graph} & \textbf{Result} & \textbf{Baseline} \\
\toprule

\cite{chen2021attention} & ABIDE (1007) [M] & 481 ASD, 526 TDC & AAL116 & New & BG & 72.7\% & GCN 70.4\% \\[.7mm]

\cite{chen2022adversarial} & ABIDE (1007) [M] & 481 ASD, 526 TDC & AAL116 & New & BG & 74.7\% & GCN 70.4\% \\

\cite{li2020graph} & Biopoint (118) & 75 ASD, 43 TDC & DX148 & GraphSAGE & BG & 70.0\% & - \\[.7mm]

\cite{li2020pooling} & Biopoint (118) & 75 ASD, 43 TDC & DK84 & GAT & BG & 79.7\% & CNN 78.1\% \\[.7mm]

\cite{li2021braingnn} & Biopoint (118) & 75 ASD, 43 TDC & DK84 & New & BG & 79.8\% & GAT 77.4\% \\[.7mm]

\cite{yang2022identification} & ABIDE (303) & 130 ASD, 173 TDC & SF200 & GIN & BG & 70.6\% & SVM 67.4\% \\[.7mm]

\cite{chu2022resting} & ABIDE (351) & 155 ASD, 196 TDC & AAL116 & Kipf & BG & 70.0\% & GCN 62.0\% \\[.7mm]

\cite{zhao2022multi} & ABIDE NYU (92) [d] & 45 ASD, 47 TDC & AAL116 & Kipf & BG & 79.9\% & FCN 72.6\% \\[.7mm]

\cite{noman2022graph} & ABIDE (144) [d] & 70 ASD, 74 TDC & Power264 & Kipf & BG & 66.0\% & SVM 63.8\% \\[.7mm]

\bottomrule
\end{tabularx}
\label{tab:supp-asd1}
\end{center}
\end{table*}

\begin{table*}[h]
\caption{Summary of findings from ASD studies, including those with biomarker discovery performed (top section), without biomarker discovery performed (middle section), and smaller datasets (bottom section). `Size' refers to the size of the dataset. When modalities beyond sFC are used, they are marked with [d] (dFC) or [M] (multimodal).
}
\begin{center}
\begin{tabularx}{\textwidth}{@{} L{0.33}L{0.65}L{0.9}L{0.4}L{0.5}L{0.50}L{0.33}L{0.67} @{}}
\toprule
\textbf{Reference} & \textbf{Dataset (Size)} & \textbf{Dataset distribution} & \textbf{Atlas} & \textbf{GNN} & \textbf{Graph} & \textbf{Result} & \textbf{Baseline} \\
\toprule

\cite{zheng2024ci} & ABIDE (1064) & 528 ASD, 536 TC & AAL116 & GCN & BG & 71.0\% & GAT 68.0\% \\
\cite{zheng2024bpi} & ABIDE (1064) & 528 ASD, 536 TC & AAL116 & GCN & BG & 71.0\% & GAT 68.0\% \\
\cite{ma2024identification} & ABIDE (714) & 334 ASD, 380 TC & AAL116 & GCN & BG & 70.3\% & BrainGNN 67.3\% \\
\cite{zheng2024brainib} & ABIDE (1099) & 528 ASD, 571 TC & AAL116 & GIN & BG & 70.2\% & GIN 67.9\% \\
\cite{wang2024multiview} & ABIDE, 4 sites (355) & 167 ASD, 188 TC & AAL116, HO111 & HGCN & BG & 78.5\% & GAT 69.2\% \\
\cite{wang2024adaptive} & ABIDE (860) & 392 ASD, 458 TC & HO110, CC200 & GCN & PG & 70.7\% & GCN 70.1\% \\
\cite{kong2024multi} & ABIDE (618) [d] & 290 ASD, 328 TC & AAL90 & GIN & BG & 72.3\% & STAGCN 69.1\% \\
\cite{fang2024bilinear} & ABIDE, 3 sites (312) [M] & 148 ASD, 164 TC & AAL116 & GCN & BG & 82.4\% & GCN 70.6\% \\

\cite{wang2024ifc} & ABIDE (871) [d] & 403 ASD, 468 TC & Unclear & Cheb & PG & 80.7\% & GCN 76.4\% \\

\bottomrule
\end{tabularx}
\label{tab:supp-asd2}
\end{center}
\end{table*}

\begin{table*}[h]
\caption{Summary of findings from ASD studies, including those with biomarker discovery performed (top section), without biomarker discovery performed (middle section), and smaller datasets (bottom section). `Size' refers to the size of the dataset. When modalities beyond sFC are used, they are marked with [d] (dFC) or [M] (multimodal).
}
\begin{center}
\begin{tabularx}{\textwidth}{@{} L{0.33}L{0.65}L{0.9}L{0.4}L{0.5}L{0.50}L{0.33}L{0.67} @{}}
\toprule
\textbf{Reference} & \textbf{Dataset (Size)} & \textbf{Dataset distribution} & \textbf{Atlas} & \textbf{GNN} & \textbf{Graph} & \textbf{Result} & \textbf{Baseline} \\
\toprule

\cite{gu2025fc} & ABIDE (871) & 403 ASD, 468 HC & HO110 & GCN & Both & 99.8\% & popGCN 96.8\% \\
\cite{wang2024leveraging} & ABIDE NYU (184) [d] & 79 ASD, 105 TC & AAL116 & GIN & BG & 73.2\% & GCN 67.5\% \\
\cite{wei2023autistic} & Private (138) [M] & 61 ASD, 77 TC & AAL90 & Cheb & BG & - & - \\
\cite{bian2024adversarially} & ABIDE (663) & 314 ASD, 349 TC & AAL90 & New & BG & 76.1\% & GCN 71.7\% \\

\cite{gu2024novel} & ABIDE (871) & 403 ASD, 468 TC & HO110 & GCN & Both & 99.8\% & popGCN 96.8\% \\

\cite{wang2023consistency} & ABIDE, 3 sites (307) & 145 ASD, 162 TC & AAL116, HO111 & HGCN & BG & 71.7\% & DNN 68.2\% \\
\cite{wang2023plsnet} & ABIDE (1009) & 570 ASD, 539 TC & AAL116, CC200 & GCN & BG & 72.4\% & FBNETGEN 71.3\% \\
\cite{zheng2023dynbraingnn} & ABIDE (582) [d] & 289 ASD, 293 TC & Unclear & GCN & BG & 75.0\% & STAGIN 72.0\% \\
\cite{xu2024contrastive} & ABIDE (989) & 455 ASD, 534 TC & SF100 & GCN & BG & 68.6\% & BrainNetCNN 65.9\% \\
\cite{zhang2023gcl} & ABIDE (987) & 467 ASD, 520 TC & Multiple & GIN & BG & 80.7\% & SAGE 71.1\% \\

\cite{zhang2023gcl} & ABIDE II (532) & 243 ASD, 289 TC & Multiple & GIN & BG & 79.9\% & SAGE 71.2\% \\
\cite{wang2023modularity} & ABIDE, NYU (184) & 79 ASD, 105 TC & AAL116 & GIN & BG & 72.6\% & GCN 67.5\% \\
\cite{menon2023asdexplainer} & Private (117) & 75 ASD, 43 TC & Unclear & GCN & BG & - & - \\

\cite{hu2021gat} & ABIDE (1035) & 505 ASD, 530 TC & HO110 & GAT & BG & 68.0\% & Cheb 63.6\% \\

\bottomrule
\end{tabularx}
\label{tab:supp-asd3}
\end{center}
\end{table*}

\begin{table*}[h]
\caption{Summary of findings from MDD studies, including those with biomarker discovery performed (placed above midline). `Size' refers to size of dataset. When modalities beyond sFC are used, they are marked with [d] (dFC) or [M] (multimodal).
}
\begin{center}
\begin{tabularx}{\textwidth}{@{} L{0.3}L{0.8}L{0.77}L{0.66}L{0.4}L{0.33}L{0.2}L{0.67} @{}}
\toprule
\textbf{Reference} & \textbf{Dataset (Size)} & \textbf{Dataset distribution} & \textbf{Atlas} & \textbf{GNN} & \textbf{Graph} & \textbf{Result} & \textbf{Baseline} \\
\toprule

\cite{qin2022using} & REST-meta-MDD (1586) & 821 MDD, 765 NC & DOS160 & Defferrard & BG & 81.5\% & - \\[3.5mm]

\cite{gallo2023functional} & REST-meta-MDD, psymri (2498) & 1249 MDD, 1249 NC & HO112, CC200 & Kipf & BG & 61.3\% & SVM-RBF 61.2\% \\[3.5mm]

\cite{kong2022multi} & Private (218) & 129 MDD, 89 NC & BM82, JHU81 & Kipf & BG & 70.9\% & GAT 68.2\% \\[3.5mm]

\cite{jun2020identifying} & Private (75) & 29 MDD, 44 NC & Yeo114 & Defferrard & PG & 74.1\% & SVM 69.8\% \\[3.5mm]

\cite{kong2021spatio} & Private (277) [d] & 180 MDD, 97 NC & Unclear & GAT & BG & 84.0\% & - \\[3.5mm]

\cite{fang2023unsupervised} & REST-meta-MDD (681) [d] & 356 MDD, 325 NC & AAL116 & Kipf & BG & 59.3\% & STNet 52.0\% \\

\cite{zheng2024ci} & REST-meta-MDD (1604) & 828 MDD, 776 HC & AAL116 & GCN & BG & 72.0\% & GAT 63.0\% \\
\cite{zheng2024bpi} & REST-meta-MDD (1604) & 828 MDD, 776 HC & AAL116 & GCN & BG & 73.0\% & GAT 63.0\% \\
\cite{lee2024spectral} & REST-meta-MDD (470) & 245 MDD, 225 HC & Multiple & Cheb & BG & 69.6\% & MMTGCN 66.9\% \\

\bottomrule
\end{tabularx}
\label{tab:supp-mdd}
\end{center}
\end{table*}

\begin{table*}[h]
\caption{Summary of findings from MDD studies, including those with biomarker discovery performed (placed above midline). `Size' refers to size of dataset. When modalities beyond sFC are used, they are marked with [d] (dFC) or [M] (multimodal).
}
\begin{center}
\begin{tabularx}{\textwidth}{@{} L{0.3}L{0.8}L{0.77}L{0.66}L{0.4}L{0.33}L{0.2}L{0.67} @{}}
\toprule
\textbf{Reference} & \textbf{Dataset (Size)} & \textbf{Dataset distribution} & \textbf{Atlas} & \textbf{GNN} & \textbf{Graph} & \textbf{Result} & \textbf{Baseline} \\
\toprule

\cite{zheng2024brainib} & REST-meta-MDD (1604) & 828 MDD, 776 HC & AAL 116 & GIN & BG & 70.0\% & GIN 65.4\% \\
\cite{kong2024multi} & Zhongda (520) [d] & 314 MDD, 206 HC & AAL 90 & GIN & BG & 73.2\% & STAGCN 69.8\% \\
\cite{kong2024multi} & REST-meta-MDD (667) [d] & 368 MDD, 299 HC & AAL90 & GIN & BG & 68.7\% & STAGCN 66.0\% \\
\cite{liu2024fusing} & REST-meta-MDD S20 (533) & 282 MDD, 251 HC & AAL116 & GCN & BG & 77.6\% & GCN 75.5\% \\
\cite{gu2025fc} & REST-meta-MDD SU (533) & 282 MDD, 251 HC & HO110 & GCN & Both & 93.4\% & popGCN 71.6\% \\
\cite{zhao2024enhancing} & REST-meta-MDD (1611) [d] & 832 MDD, 779 HC & AAL116 & GIN & BG & 67.1\% & STAGIN 64.1\% \\
\cite{gu2024novel} & REST-meta-MDD SU (533) & 282 MDD, 251 HC & HO110 & GCN & Both & 93.4\% & popGCN 71.6\% \\
\cite{zhang2023slg} & REST-meta-MDD (2361) & 1256 MDD, 1105 HC & AAL116 & GCN & BG & 74.2\% & GCN (PG) 61.0\% \\
\cite{kong2023multi} & Private (187) [M] & 93 MDD, 94 HC & AAL116, Destrieux, HO & GCN & BG & 80.2\% & MMGNN 73.6\% \\
\cite{dai2023classification} & REST-meta-MDD (615) & 189 MDD, 426 HC & AAL116, CC200, DOS & GCN & Temporal & 75.8\% & 70.7\% \\
\cite{zhao2022detecting} & REST-meta-MDD (2361) & 1256 MDD, 1105 HC & AAL116 & GIN & BG & 64.8\% & SAGE 61.5\% \\
\cite{wang2023modularity} & REST-meta-MDD 20 (533) [d] & 282 MDD, 251 HC & AAL116 & GIN & BG & 60.9\% & GCN 55.7\% \\

\bottomrule
\end{tabularx}
\label{tab:supp-mdd2}
\end{center}
\end{table*}

\begin{table*}[h]
\caption{Summary of findings from SZ studies, including those with biomarker discovery performed (placed above midline). `Size' refers to the size of the dataset. When modalities beyond sFC are used, they are marked with [d] (dFC) or [M] (multimodal).
}
\begin{center}
\begin{tabularx}{\textwidth}{@{} L{0.33}L{0.6}L{0.92}L{0.7}L{0.37}L{0.50}L{0.33}L{0.5} @{}}
\toprule
\textbf{Reference} & \textbf{Dataset (Size)} & \textbf{Dataset distribution} & \textbf{Atlas} & \textbf{GNN} & \textbf{Graph} & \textbf{Result} & \textbf{Baseline} \\
\toprule

\cite{lei2022graph} & Private (1412) & 505 SZ, 907 NC & AAL116, DOS160 & Defferrard & BG & 85.8\% & SVM 80.9\% \\[3.5mm]

\cite{chen2023discriminative} & Private (345) [M] & 140 SZ, 205 NC & AAL90, BNA246 & Kipf & BG & 95.8\% & SVM 81.2\% \\[3.5mm]

\cite{sebenius2021multimodal} & COBRE (154) [M] & 67 SZ, 87 NC & DK293 & Kipf & BG & 75.0\% & SVM 71.0\% \\

\cite{zheng2024ci} & SRPBS (184) & 92 SZ, 92 HC & AAL116 & GCN & BG & 93.0\% & GAT 84.0\% \\
\cite{zheng2024bpi} & SRPBS (184) & 92 SZ, 92 HC & AAL116 & GCN & BG & 91.0\% & GAT 84.0\% \\
\cite{zhu2024temporal} & COBRE (112) [d] & 48 SZ, 64 HC & AAL90 & New & BG & 83.6\% & BrainGNN 81.3\% \\
\cite{zhu2024temporal} & UCLA (80) [d] & 41 SZ, 39 HC & AAL90 & New & BG & 89.7\% & BrainGNN 84.9\% \\
\cite{wang2024optimizing} & SRPBS (647) & 142 SZ, 505 HC & BV140 & GCN & BG & 82.5\% & GAT 75.4\% \\
\cite{sunil2024graph} & UCLA (177) & 50 SZ, 122 HC & Custom (AAL3 + HO Cortical) & GCN & BG & 82.0\% & GCN 77.0\% \\
\cite{li2023classification} & In-house (143) [M] & 70 SZ, 73 HC & AAL116 & GCN & BG & 78.4\% & SAGE 75.1\% \\
\cite{fan2023dgst} & Multiple (1003) [d] & 411 SZ, 592 HC & AAL116 & GIN & BG & 82.8\% & STAGIN 71.5\% \\

\bottomrule
\end{tabularx}
\label{tab:supp-sz}
\end{center}
\end{table*}

\end{document}